\documentclass[10pt,twocolumn,letterpaper]{article}

\usepackage[pagenumbers]{iccv} 

\usepackage{multirow}

\usepackage{multirow}
\usepackage{makecell}
\usepackage{array}
\usepackage{caption}
\usepackage{multirow}
\usepackage{ragged2e}
\usepackage{colortbl}
\usepackage{pifont}
\usepackage{bbding}
\usepackage[numbers,sort&compress]{natbib}
\usepackage{listings}
\usepackage{tcolorbox}
\definecolor{iccvblue}{rgb}{0.21,0.49,0.74}
\usepackage[pagebackref,breaklinks,colorlinks,allcolors=iccvblue]{hyperref}

\usepackage{url}
\usepackage{hyperref}

\def\paperID{10221} 
\def\confName{ICCV}
\def\confYear{2025}

\title{CoLMDriver: LLM-based Negotiation Benefits \\ Cooperative Autonomous Driving}

\author{Changxing Liu\textsuperscript{1,3\thanks{Equal contribution}} \quad Genjia Liu\textsuperscript{1,3\footnotemark[1]} \quad Zijun Wang\textsuperscript{1,3} \quad Jinchang Yang\textsuperscript{1,3} \quad Siheng Chen\textsuperscript{1,2,3\thanks{Corresponding author}}
\\
\\
\textsuperscript{1}Shanghai Jiao Tong University \quad \textsuperscript{2}Shanghai AI Laboratory \\
\textsuperscript{3}Multi-Agent Governance \& Intelligence Crew (MAGIC) \quad 
\\
{\tt\small \{cx-liu,LGJ1zed,wzjinsjtu,andreo\_y,sihengc\}@sjtu.edu.cn}
}

\begin{document}

\maketitle

\begin{abstract}

Vehicle-to-vehicle (V2V) cooperative autonomous driving holds great promise for improving safety by addressing the perception and prediction uncertainties inherent in single-agent systems. However, traditional cooperative methods are constrained by rigid collaboration protocols and limited generalization to unseen interactive scenarios. While LLM-based approaches offer generalized reasoning capabilities, their challenges in spatial planning and unstable inference latency hinder their direct application in cooperative driving. To address these limitations, we propose \textbf{CoLMDriver}, the first full-pipeline LLM-based cooperative driving system, enabling effective language-based negotiation and real-time driving control. CoLMDriver features a parallel driving pipeline with two key components: (i) an LLM-based negotiation module under an actor-critic paradigm, which continuously refines cooperation policies through feedback from previous decisions of all vehicles; and (ii) an intention-guided waypoint generator, which translates negotiation outcomes into executable waypoints. Additionally, we introduce \textbf{InterDrive}, a CARLA-based simulation benchmark comprising 10 challenging interactive driving scenarios for evaluating V2V cooperation. Experimental results demonstrate that CoLMDriver significantly outperforms existing approaches, achieving an 11\% higher success rate across diverse highly interactive V2V driving scenarios. Code will be released on \url{https://github.com/cxliu0314/CoLMDriver}.

\end{abstract}    
\vspace{-3mm}
\section{Introduction}
\label{sec:intro}

Vehicle-to-vehicle (V2V) cooperative autonomous driving (AD) aims to improve driving performance by allowing autonomous vehicles to communicate with surrounding vehicles.
Unlike single-vehicle autonomous driving~\cite{InterFuser,TransFuser,UniAD,jiang2023vad,shao2023reasonnet}, where each vehicle makes driving decisions based solely on the observations from its own sensors, cooperative driving enables vehicles to exchange driving-related data~\cite{zhang2024MARL_survey,han2023coperception_survey}. This collaborative information-sharing mechanism helps autonomous vehicles surmount the inherent limitations in single-vehicle driving, such as incomplete environmental perception~\cite{liu2024codriving,HuWhere2comm:NeurIPS22,lu-heal-2023,wei-cobevflow-2023} and uncertainty in forecasting the future states of surrounding traffic participants~\cite{liu2024ADMM,chen2022conflict}.


\begin{figure}[!t]
  \centering
  \includegraphics[width=0.47\textwidth]{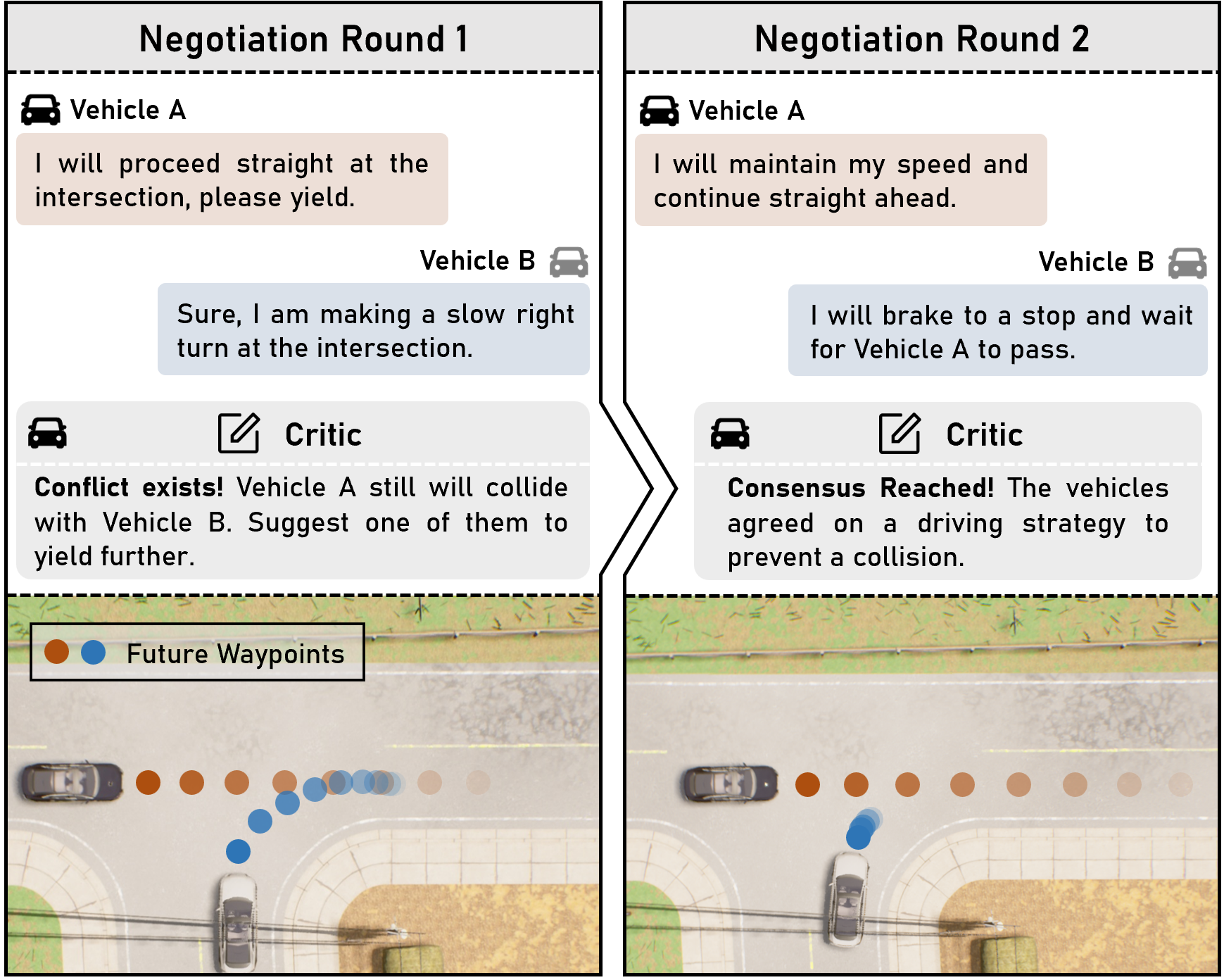}
  \caption{Negotiation with critic feedback. CoLMDriver refines cooperation policy by evaluating the latest negotiation outcomes.}
  \vspace{-4mm}
  \label{fig:feedback_example}
\end{figure}

Traditional cooperative driving approaches can be generally categorized into optimization-based and learning-based methods. 
Optimization-based cooperative driving methods~\cite{liu2024ADMM,huang2023decentralized,huang2023parallel} formulate multi-vehicle planning as constrained optimization problems to determine optimal actions. However, these methods depend on precise environmental modeling and require task-specific optimization objectives and constraints, making them inherently limited in handling unknown scenarios.
Learning-based methods~\cite{zhao2023multi,zheng2024safe,zhan2019interaction_dataset} employ reinforcement learning and imitation learning to develop cooperative driving policies. While these approaches have been applied to several driving tasks~\cite{zhou2022cooperative,liu2024cooperative,chen2024communication}, they struggle with declined performance when encountering unseen multi-vehicle interaction patterns~\cite{kirk2023survey,ghosh2021generalization}. These limitations underscore the exploration towards a more flexible and generalizable cooperative driving framework.


Recently, large language models (LLMs) have gained significant attention in cooperative systems~\cite{li2024survey,han2024llm} due to their remarkable reasoning abilities and vast knowledge. This advancement underscores the potential of LLM-based cooperative driving, where vehicles negotiate through natural language.
Compared to optimization-based and learning-based approaches, LLM-based cooperation offers two key advantages. 
First, language-based cooperation offers greater flexibility compared to fixed-protocol communication~\cite{liu2024ADMM}, as it can incorporate both local motion details and global scene semantics. 
Second, with extensive pre-trained commonsense knowledge, LLMs have demonstrated strong capabilities in understanding traffic scenarios and making driving decisions~\cite{wen2023dilu,shao2024lmdrive,sima2023drivelm}. This indicates their potential to handle diverse multi-vehicle driving scenarios, including complex cases such as navigating non-traffic-light intersections.
However, integrating LLMs into cooperative driving faces three challenges. First, LLMs' limited ability to understand and plan in continuous road spaces makes direct application infeasible~\cite{mao2023gpt}, requiring additional spatial information for effective cooperation. Second, redundant environmental information and unconstrained negotiation reduce efficiency, necessitating selective communication with relevant collaborators. Third, LLMs’ long and unstable inference delays hinder high-frequency planning, demanding efficient negotiation and inference mechanisms to adapt to real-time control.

To address these challenges, we propose \textbf{CoLMDriver} (Cooperative Language-Model-based Driver), the~\emph{first} full pipeline (from sensor data to control signal) LLM-based cooperative driving system that accommodates  real-time control with efficient planning negotiation.
CoLMDriver consists of two parallel planning pipelines: i)~\emph{An end-to-end driving pipeline} for real-time vehicle control, inherently capable of full driving functionality. To integrate language-based negotiation, we incorporate it with an intention-guided waypoint planner that translates high-level negotiation outcomes into executable waypoints. ii)~\emph{A cooperative planning pipeline}, implemented via an LLM-based negotiation module. To enhance the effectiveness and efficiency of negotiation, we propose three key techniques. First, we introduce an Actor-Critic feedback mechanism that evaluates negotiation outcomes and feeds the results back to the LLM-based negotiator, enabling continuous policy refinement as shown in Fig~\ref{fig:feedback_example}. This evaluation considers both high-level intentions and low-level waypoints, providing feedback from safety, efficiency, and multi-vehicle consensus perspectives. Second, we propose a dynamic grouping mechanism to select relevant collaborators for negotiation, improving efficiency by focusing on critical agents. Third, we integrate an auxiliary VLM-based intention planner to handle non-cooperation periods.

This system offers two key advantages. First, it effectively integrates LLM-based cooperative planning with fine-grained waypoint generation. LLM-derived driving intentions guide waypoints generation, and the waypoints provide feedback to refine cooperation strategies, forming an online optimization loop. Second, its parallel framework accommodates asynchronous planning, mitigating the inherent inference latency gap between the LLM and the end-to-end pipeline.



To evaluate performance in V2V scenarios, we introduce InterDrive (Interactive Driving) benchmark, which constructs 10 challenging traffic scenarios in the CARLA simulator~\cite{DosovitskiyCARLA:CoRL2017}. These scenarios involve multiple autonomous vehicles with severely conflicting routes, testing an AD system’s ability to handle highly interactive V2V situations.
We evaluate CoLMDriver on both InterDrive benchmark and the public Town05 benchmark~\cite{TransFuser}. Results indicate that CoLMDriver surpasses existing single-vehicle and cooperative driving methods, achieving an 11\% higher success rate across diverse scenarios.

To sum up, our contributions are:
\begin{itemize}
    \item We propose CoLMDriver, the first full pipeline LLM-based cooperative driving system, featuring two main components: an LLM-based negotiator with Actor-Critic feedback, and an intention-guided waypoints planner to translate negotiation outcomes.
    \item We introduce InterDrive Benchmark, which includes 10 types of challenging scenarios to enable the evaluation of autonomous driving in handling V2V interactions.
    \item We conduct comprehensive experiments and validate that CoLMDriver achieves a superior success rate in various V2V driving scenarios.
\end{itemize}
\vspace{-3mm}
\section{Related works}
\label{sec/2_related_works}

\subsection{End-to-end Autonomous Driving}
A key research direction in end-to-end autonomous driving is imitation learning, which aims to replicate expert driving behaviors by fitting a model to recorded driving data. Recent advancements focus on several core areas to improve driving performance and robustness. Some methods, such as NEAT~\cite{NEAT}, TransFuser~\cite{TransFuser}, UniAD~\cite{UniAD}, InterFuser~\cite{InterFuser}, and ReasonNet~\cite{shao2023reasonnet}, leverage transformer architectures to capture more nuanced representations of driving scenarios, enhancing the model’s ability to process complex environments. Other approaches, like MP3~\cite{MP3}, UniAD, LAV~\cite{LAV}, TCP~\cite{TCP}, incorporate auxiliary tasks that provide additional learning signals to support the primary driving task, leading to better generalization.
However, the IL approach struggles with low generalization to unseen scenarios and lacks causal reasoning. To overcome these issues, we propose an LLM-based method to achieve generalized reasoning ability in diverse interactive scenarios.


\subsection{MLLMs-based Driving}

In the field of autonomous driving (AD), recent research~\cite{mao2023language,wen2023dilu,hu2024agentscodriver,sha2023languagempc} has integrated LLMs into AD systems to improve interpretability and facilitate human-like interactions. Some studies~\cite{renz2024carllava,wang2023drivemlm,tian2024drivevlm,sima2023drivelm} leverage VLMs to process multi-modal input data, providing both descriptive text and control signals suited to driving scenarios. LMDrive~\cite{shao2024lmdrive} integrates multi-modal sensor data with textual instructions, leveraging LLMs for closed-loop end-to-end AD. 

Most current research focuses on using LLMs to enhance individual driving capabilities, while a few works explore driving cooperation.
AgentsCoDriver~\cite{hu2024agentscodriver} promotes lifelong learning through interaction with the environment, enabling simple negotiations between agents. CoDrivingLLM~\cite{fang2024towards} centered around roadside units for vehicle-to-vehicle negotiations to resolve conflicts. 
However, these approaches are limited to discrete decisions and cannot generate executable control signals. They also overlook the inference latency of LLM, making real-world deployment challenging.
To bridge these gaps, we propose CoLMDriver, an LLM-based cooperative system that generates real-time driving signals through a parallel framework.

\vspace{-2mm}
\section{Problem Formulation}
\label{subsec:problem_formulation}
Consider $N$ agents participate in the cooperation. Let $\mathcal{X}_i$ and $\mathcal{D}_i$ be the observation and the destination of the $i$th agent. The objective of collaborative driving is to achieve the maximized driving performance of all agents; that is, 
\vspace{-2mm}
\begin{equation}
    \arg\max_{\theta,\mathcal{M}}\sum_{i=1}^N  d(\Phi_\theta(\mathcal{X}_i, \mathcal{D}_i, \mathcal{M}_i^k))
\end{equation}

\vspace{-2mm}
\noindent where $d(\cdot)$ is the driving performance metric, $\Phi$ is the driving framework with trainable parameter $\theta$. $\mathcal{M}_i$ is the message exchange between agent $i$ and other agents, which can iterate $k$ rounds. Here we focus on leveraging the flexibility of language to achieve planning consensus and improve overall performance, where $\mathcal{M}_i^k = [\{\mathcal{M}_{i\leftrightarrow j}\}_{j=1}^N]^k$ represents a multi-round language-based negotiation process.

\vspace{-2mm}
\section{Methodology}

\begin{figure*}[!t]
  \centering
  \includegraphics[width=0.94\textwidth]{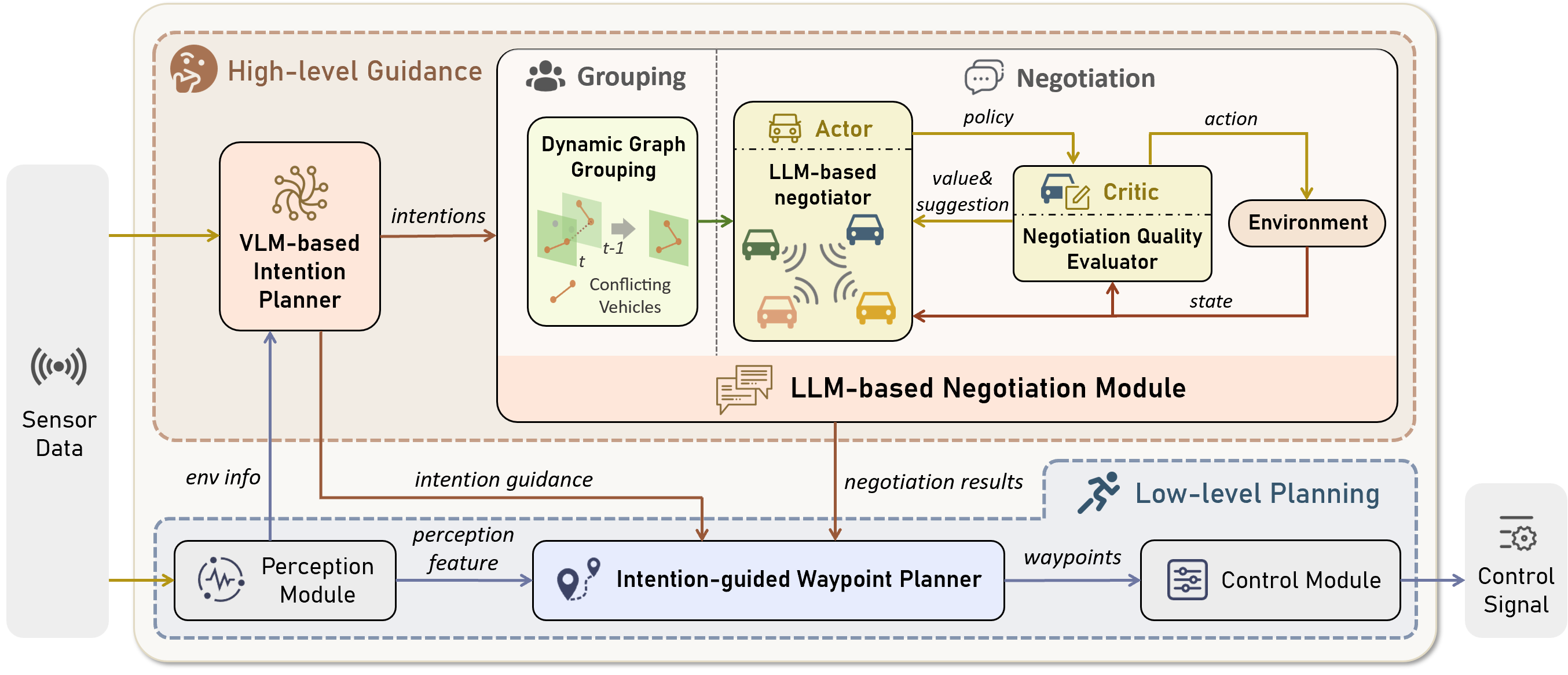}
  \caption{Overall architecture of CoLMDriver. CoLMDriver operates through a parallel driving pipeline, where language negotiation assists in the planning process through asynchronous connection of three component: i) an LLM-based negotiation module under the Actor-Critic paradigm; ii) a VLM-based intention planner and iii) an intention-guided waypoint planner.}
  \vspace{-4mm}
  \label{fig:pipeline}
\end{figure*}

This section introduces CoLMDriver, a cooperative driving system that leverages language-based negotiation and planning to enhance the collective driving capabilities of multiple autonomous vehicles. We start by outline the overall system architecture in Sec.~\ref{subsec:system_outline}, followed by detailed composition of two parallel pipelines in Sec.~\ref{subsec:high}, ~\ref{subsec:low}.

\subsection{Overall Architecture}
\label{subsec:system_outline}
As illustrated in Fig.~\ref{fig:pipeline}, CoLMDriver operates through a parallel driving pipeline designed to tackle the latency challenges of negotiation without disrupting the normal execution of the downstream planner. The high-level guidance generation pipeline conducts deep reasoning at a relatively low frequency to formulate comprehensive and consensus-driven driving intentions, while the low-level perception-planning-control pipeline runs at high frequency to ensure real-time vehicle control.

The high-level pipeline orchestrates cooperative decision-making through two core components: i) a LLM-based negotiation module under the Actor-Critic paradigm, where LLMs enable multi-round negotiation between vehicles to reach a consensus on driving policy, guided by feedback from the evaluator; ii) a VLM-based intention planner, which generate high-level driving intentions by synthesizing multi-modal environmental context. The VLM intention planner continuously refines driving intentions based on textual descriptions of the current state, detected objects from the low-level perception module and the front camera input. If conflicts are predicted, the LLM negotiation module first conduct dynamic graph grouping with surrounding vehicles to form negotiation groups, and then takes current driving intention and engages in a multi-round negotiation process with guidance from evaluator. The negotiation results and intention guidance are then fed back into the low-level waypoint planner to guide precise planning. 

The low-level pipeline follows the perception-planning-control structure. When receiving the sensor data, the perception module generates object-level 3D information and BEV perception features, conducting spatial understanding as auxiliary inputs for planning tasks. To translate language-based information into actionable waypoints, the key component is the intention-guided waypoint planner, which leverages both perception features and high-level planning intentions to generate waypoints. These waypoints are converted into control signals by the control module, resulting in improved cooperative driving outcomes.

\subsection{High-level Guidance Layer}
\label{subsec:high}
The high-level guidance pipeline is responsible for strategic decision-making and cooperative negotiation, enhancing driving adaptability through semantic reasoning and multi-agent consensus. It consists of two core components: the VLM-based intention planner and the LLM-based negotiation module. The negotiation results guide the low-level planner during the negotiation process, while the VLM output takes precedence when no negotiation is activated.

\subsubsection{LLM-based Negotiation Module}
\label{subsec:negotiator}
The LLM-based negotiation module engages in multiple rounds of dialogue with surrounding intelligent vehicles, resolving predicted conflicts by reaching a consensus on driving policies. Given the negligible latency in LLM inference, the negotiation system focuses on how to efficiently achieve consensus on a optimized driving policy. To ensure the generalizability of negotiations, we avoid imposing strict output formats or rigid communication rules. However, overly unrestricted negotiations may struggle to converge on a consensus. The key innovation lies in incorporating an \textbf{Actor-Critic paradigm} within the negotiation system. 
The Actor-Critic paradigm is a reinforcement learning approach where the "actor" selects actions based on current policies, while the "critic" evaluates the chosen actions by providing feedback on their quality, enabling faster convergence towards optimal outcomes. In our method, the LLM-based negotiators act as the actors and the evaluator as the critic. By providing feedback based on dialogue quality, safety, and efficiency expectations, we leverage the in-context learning capabilities of LLMs to facilitate rapid convergence in the negotiation process.
The LLM-based negotiation module consists of three main components: i) The dynamic graph grouping mechanism, which identifies agents with negotiation needs and establishes communication in dynamic traffic scenarios, ii) The LLM-based negotiator, which conducts negotiations with grouped agents using natural language and iii) The negotiation quality evaluator, which acts as a critic, providing feedback to the negotiator to accelerate consensus achievement. 

\noindent
\textbf{Overall Process. }
Once the negotiation process begins, vehicles first form negotiation groups using a dynamic graph grouping mechanism. In each round, vehicles take turns "speaking" in a designated order. The negotiation quality evaluator then assesses the situation, providing feedback on consensus, safety, and efficiency. The LLM-based negotiators incorporate this feedback into their input, adjust their driving intentions accordingly, and call the evaluator again. After several rounds, when the evaluator determines that consensus has been reached, the negotiation concludes, and the final driving intentions are passed on to the downstream planners of each vehicle.

\noindent
\textbf{Dynamic Graph Grouping Mechanism. }
It is crucial for vehicles to determine who and when to communicate with. To address this challenge, we prioritize vehicle groups that are most likely to conflict and build communication graph to promote effective negotiation. We assume that vehicles can automatically establish communication within the range of their hardware and are capable of broadcasting essential information, such as their planned future waypoints.

To better clarify the mutual influence between vehicles, we conduct dynamic grouping by constructing a spatiotemporal vehicle graph. Each vehicle is treated as a node, and vehicles that could potentially conflict in the future are connected by edges, which are calculated based on their safety scores derived from their waypoints. At any given moment, we build the spatial vehicle graph and apply Depth-First Search (DFS) to gather all connected vehicles into groups. To avoid inconsistent driving policies due to the constantly changing nature of dynamic groups, we preserve historical groups and merge intersecting groups across temporal dimension, obtaining a comprehensive grouping result. The communication graph $\mathcal{G}$ at time $T$ is constructed iterative:
\begin{equation}
    \mathcal{G} = \mathcal{H}^T,\quad \mathcal{H}^t = \Phi(\mathcal{H}^{t-1} \cup \mathcal{C}^t)
\end{equation}
\begin{equation}
    \mathcal{C}^t = \bigcup_k \text{DFS}(V^t, \{(v_i,v_j) \mid S_s(v_i,v_j)\geq\theta\})
\end{equation}
where safety score $S_s$ determines edges and $\Phi(\cdot)$ merges all groups that intersect between history group $\mathcal{H}^{t-1}$ and current group $\mathcal{C}^t$. Negotiations are then carried out within each group, allowing for local optimization of driving policies, which contributes to improved overall performance.

\noindent
\textbf{LLM-based negotiator. }
The LLM-based negotiator conducts human-like language negotiation with other vehicles in the group. Inputs include ego vehicle's current speed, intention, other cars' broadcast information, history conversation and critic's suggestion if exist. Since the inference time of an LLM is proportional to the output length, we have carefully designed the prompts to ensure concise information transmission and employed prompt caching techniques to maintain timeliness. The LLM-based negotiator integrates the shared information from group members, consider past conversations, and combine feedback from evaluators to output information that may include self actions, requests or responses to others. In a group that has $n$ vehicles, the negotiator output of the $i_{th}$ vehicle at round $K$ is:
\vspace{-2mm}
\begin{equation}
    O_{\text{LLM}_i^K} = LLM_i(f_{\text{P}}(\bigcup_{j=0}^n I_j, \bigcup_{k=0}^K\bigcup_{j=0}^n O_{\text{LLM}_j^k}, S^{K-1})
\end{equation}

\vspace{-2mm}
where $I$ denotes the current information shared by vehicles in the group, including speed, intention, and position, and $S$ the evaluator's suggestion, $f_{\text{P}}$ the prompt generation process. Since the used LLM is not trained on a specific domain, this paradigm differs from previous multi-vehicle cooperative driving approaches by not requiring each vehicle to be equipped with a specific model, demonstrating the versatility and broad applicability of LLM.

\noindent
\textbf{Negotiation Quality Evaluator. }
The negotiation quality evaluator acts as a critic, assessing the negotiation performance based on future planning and generating feedback related to consensus, safety, and efficiency concerns. The evaluation process follows three key steps: sum, score, and criticize. To initiate the evaluation, the evaluator can be activated on a random vehicle within the group. Based on the current round conversation, the evaluator first sums each vehicle's actions using LLM, transforming them into driving intention formats, and then distributes the results to all vehicles. Each vehicle's waypoint planner uses the summed intentions as input, generates planned waypoints, and broadcasts these plans to assist in the evaluation. The evaluator conducts the scoring process by assessing three key aspects — consensus, safety, and efficiency. Consensus score $S_c$ is judged by LLM, indicating whether every vehicle in the group is willing to execute the reached policy. Both safety score $S_s$ and efficiency score $S_e$ are derived from the waypoints, calculated by the carefully designed formula:
\begin{equation}
    S_c^k = LLM_c(\bigcup_{j=0}^n O_{\text{LLM}_j^k}),[S_s^k, S_e^k] = \mathcal{F}(\bigcup_{j=0}^n W_j)
\end{equation}
where $W$ is the predicted waypoints and $\mathcal{F}$ the score calculation formula. Finally, the evaluator provides feedback $\mathcal{R}$ through a classifier $\Psi$, criticizing scores that fail to meet the required standards. 
\begin{equation}
    \mathcal{R} = \Psi(S_c^k,S_s^k,S_e^k)
\end{equation}
This criticism is used as input for the next round of negotiation, guiding the system towards an optimal driving policy by encouraging faster convergence.

\subsubsection{VLM-based Intention Planner}
\label{subsec:intention_planner}
The VLM-based intention planner utilizes the generalized knowledge embedded in language models to recognize unusual objects and deal with complex scenes, providing more holistic decision support. The focus is to provide optimal high-level driving intention to accurately guide the downstream planner. To comprehensively and efficiently activate the understanding and decision-making capabilities of the VLM-based intention planner, we have carefully designed a hierarchical prompt generation process and limited output format. The prompt contains perception results written in an intelligible format, providing accurate environment information. To collect reasonable driving intention in different environments, we use V2Xverse~\cite{liu2024codriving} platform and employ an expert agent~\cite{InterFuser} to record driving data, capturing a wide range of urban scenarios. Driving intentions are defined as navigation intentions and speed intentions. Navigation intentions are derived from the ground truth navigation instructions, while speed intentions are extracted from the expert’s driving speed. To adapt the VLM to the specific task of driving intention assessment, we utilize the processed driving data for transfer learning based on LoRA.

\subsection{Low-level Planning Layer}
\label{subsec:low}
The low-level planning layer focuses on real-time execution, translating high-level intentions into geometrically feasible trajectories and control commands. The key component is the intention-guided waypoint planner, operating at high frequency to conduct precise planning guided by driving intentions.

\subsubsection{Intention-guided Waypoint Planner}
\label{subsec:waypoints_translator}
The Intention-guided waypoint planner acts a bridge connecting high-level driving intentions and low-level implementation paths. The challenge lies in how to precisely map high-level intentions to specific scenarios as usable waypoints. Our design consists of two main parts: intention-to-waypoint data generation and the model structure.

\noindent
\textbf{Intention-to-waypoint Data Generation. }To achieve precise intention-guided waypoints generation, we use waypoints of expert agent as a reference and generate waypoints that align with the intended action while satisfying practical scenario constraints. Based on the observation that acceleration is influenced by surrounding objects density, we extract the actual waypoints of the referenced vehicle and interpolate them using an environment-adaptive acceleration model, which generates elaborate waypoints corresponding to different driving intentions. Given a ground-truth waypoints $W$, the data generation process can be expressed as $W_g = \Phi(W, a)$. Here, the acceleration $a=f(I, x, \sigma)$ is guided by the intention $I$ and generated by the environment-adaptive acceleration model $f$, considering the distance $x$ to the nearest vehicle and the vehicle density $\sigma$. The generated waypoints $W_g$ is interpolated by function $\Phi$, which conforms to driving norms and adapts to environmental conditions.

\begin{figure}[!t]
  \centering
  \includegraphics[width=0.45\textwidth]{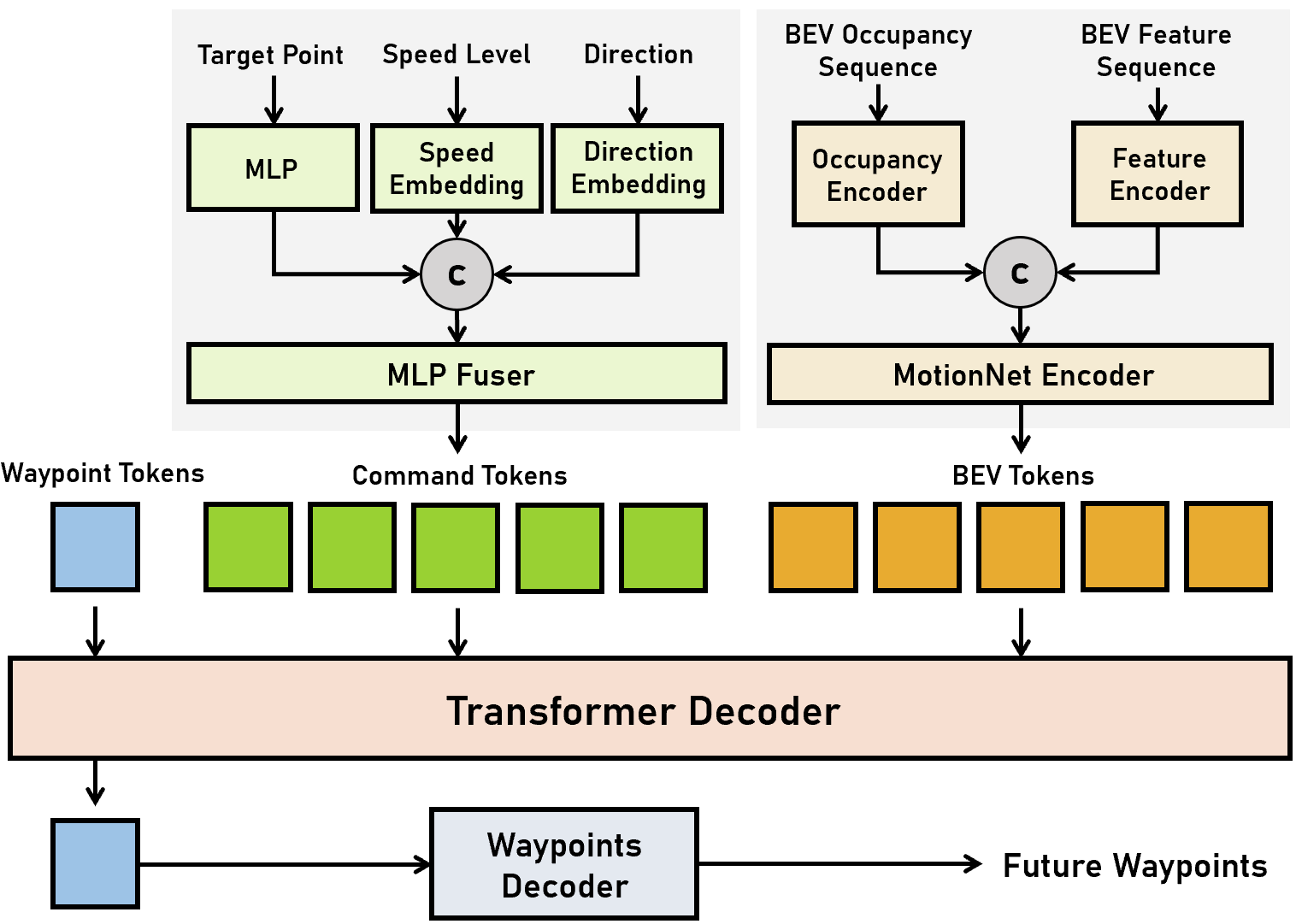}
  \caption{Model architecture of the low-level Transformer-based intention-guided waypoint planner. }
  \vspace{-4mm}
  \label{fig:generator}
\end{figure}

\noindent
\textbf{Model Structure. }To ensure waypoints align with different driving intentions within the same scenario, we developed a Transformer-based, intention-guided waypoint planner, as shown in Fig.~\ref{fig:generator}. The model effectively takes input from the BEV occupancy map and BEV features from previous frames, which are processed by the MotionNet~\cite{wu2020motionnet} encoder to capture the environmental context. Additionally, goal-oriented inputs, including target points, navigation intentions, and speed intentions, are fused through a MLP Fuser to form the guidance context. A multi-layer Transformer decoder performs cross-attention between a waypoint query and the environmental/guidance contexts, followed by a Waypoints Decoder to generate a sequence of waypoints. These waypoints are then passed to the control module to produce the necessary control signals.

\section{InterDrive Test Benchmark}

\begin{figure*}[!t]
    \captionsetup{justification=raggedright}
    \centering
    \begin{subfigure}{0.19\textwidth}  
        \centering
        \includegraphics[width=\linewidth]{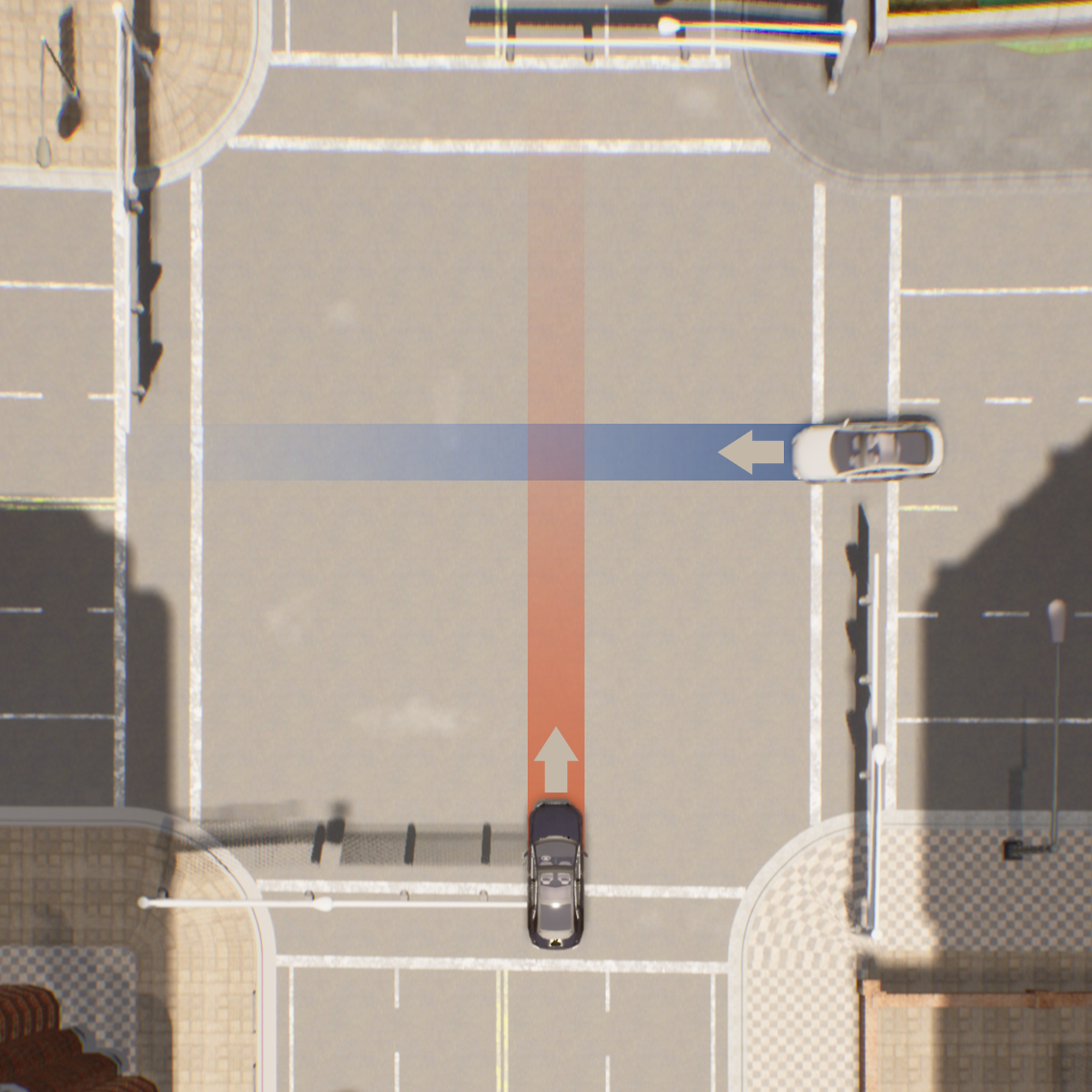}  
        \caption{Cross:Straight-Straight}
        \label{fig:scen_1}
    \end{subfigure}
    \begin{subfigure}{0.19\textwidth}
        \centering
        \includegraphics[width=\linewidth]{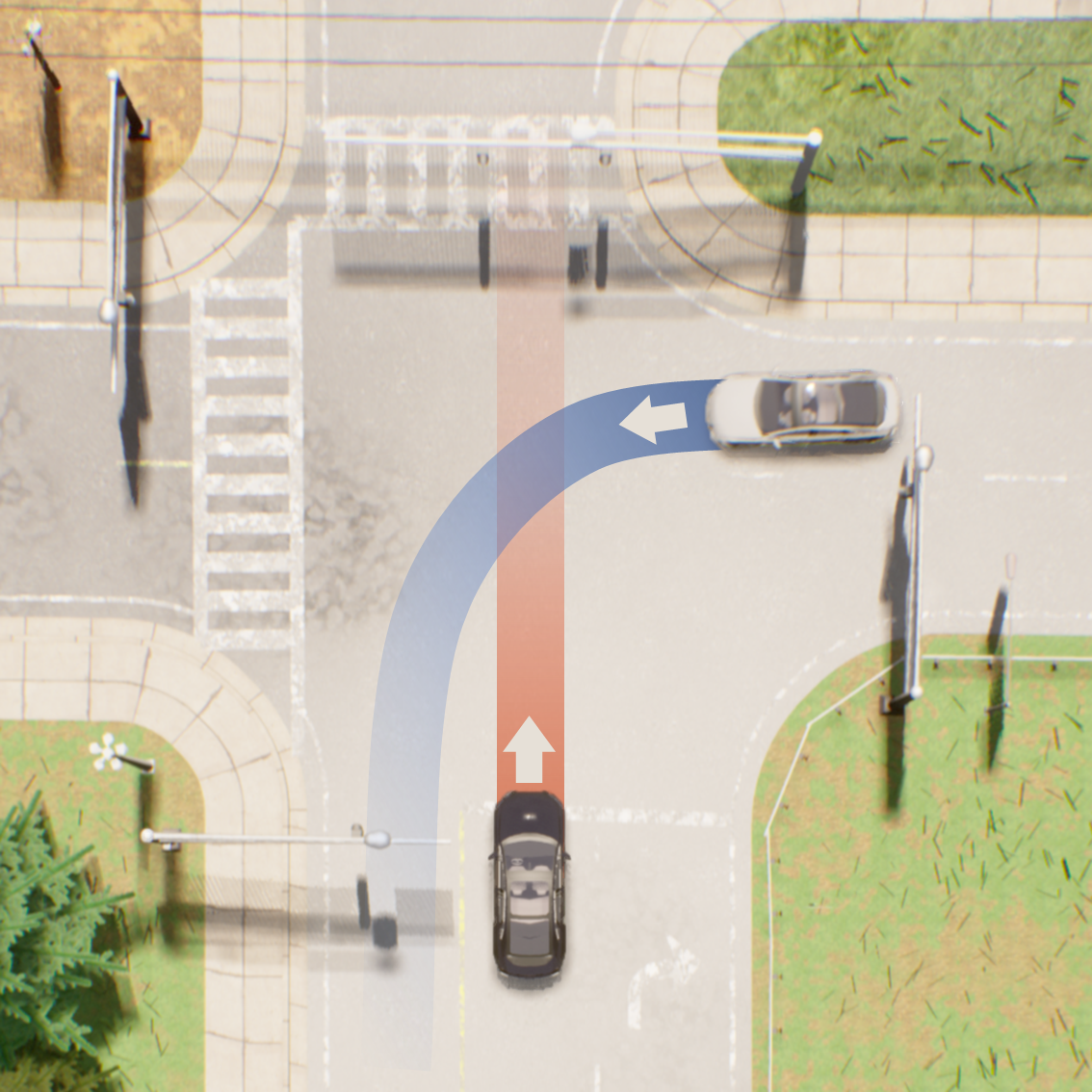}
        \caption{Cross:Straight-Left}
        \label{fig:scen_2}
    \end{subfigure}
    \begin{subfigure}{0.19\textwidth}
        \centering
        \includegraphics[width=\linewidth]{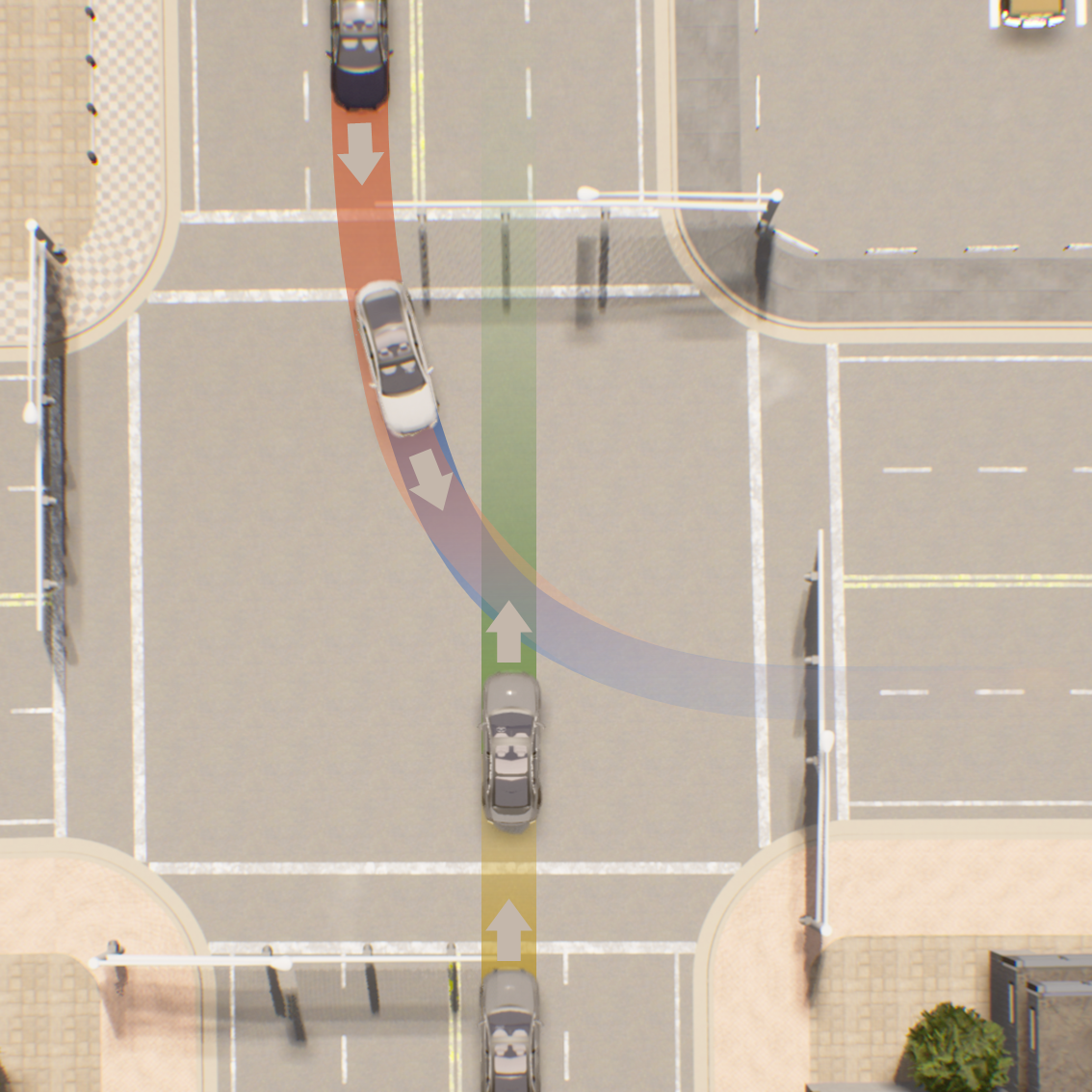}
        \caption{Cross:Opposite Lane}
        \label{fig:scen_3}
    \end{subfigure}
    \begin{subfigure}{0.19\textwidth}
        \centering
        \includegraphics[width=\linewidth]{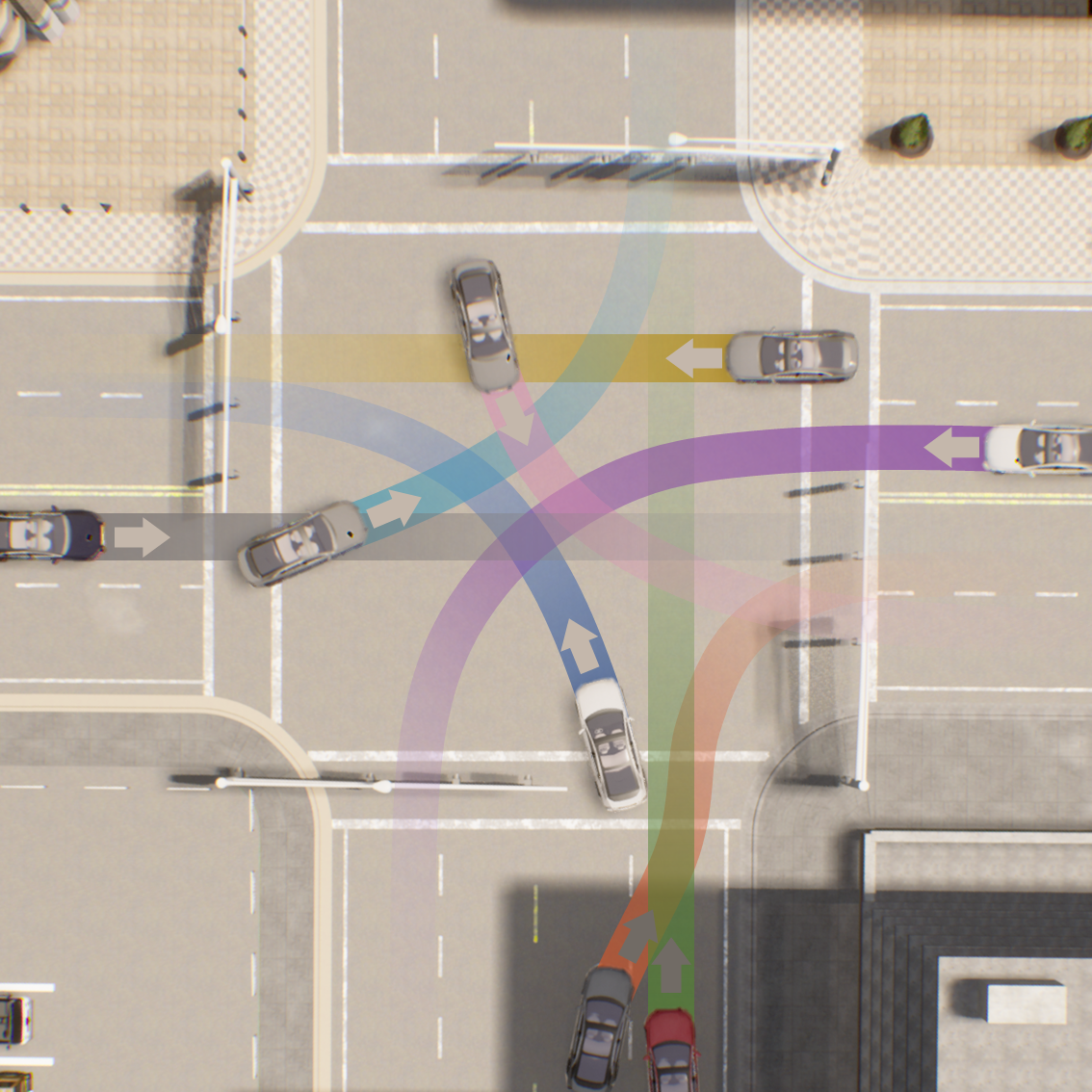}
        \caption{Cross:Chaos}
        \label{scen_4}
    \end{subfigure}  
    \begin{subfigure}{0.19\textwidth}
        \centering
        \includegraphics[width=\linewidth]{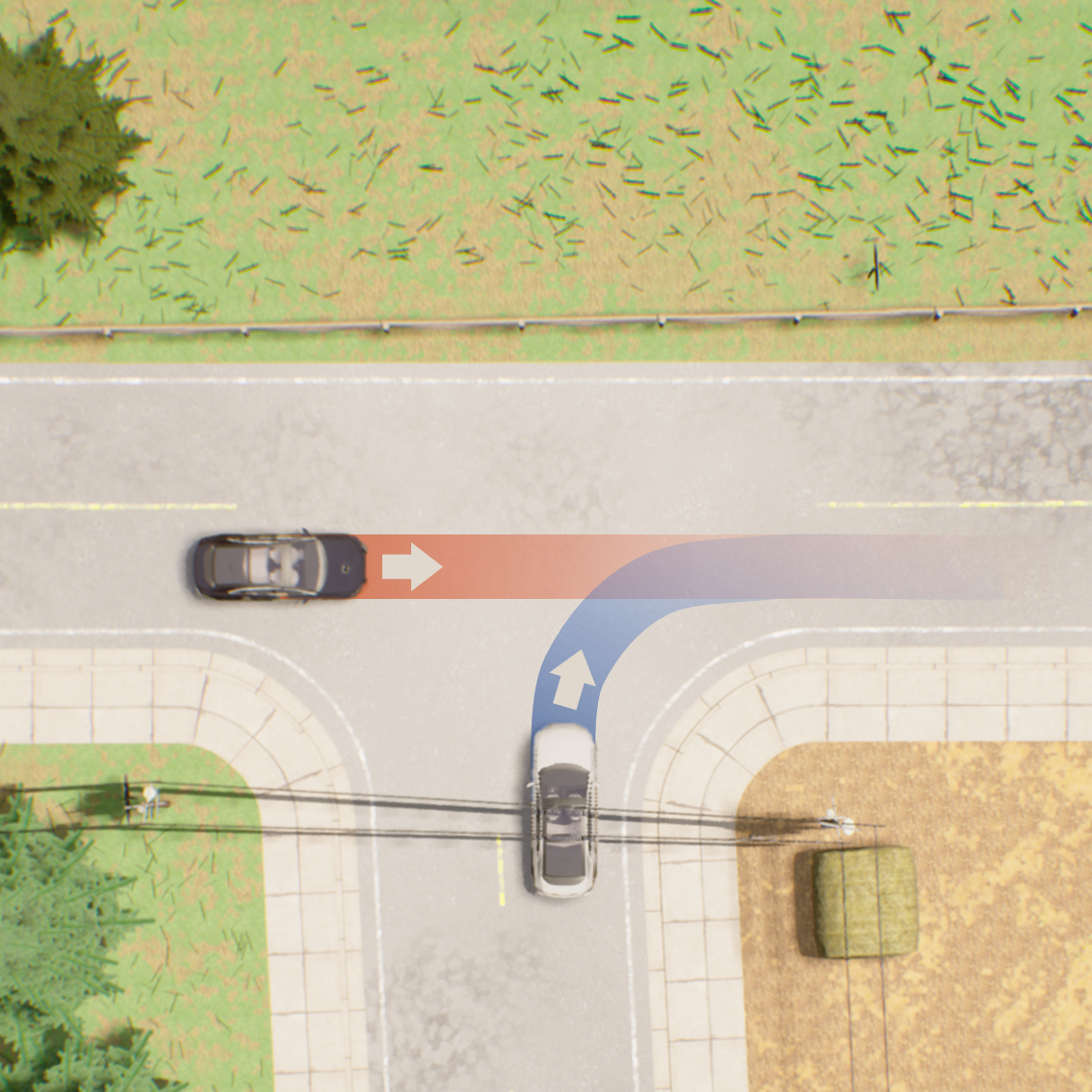}
        \caption{Merge:Straight-Right}
        \label{scen_5}
    \end{subfigure}  
    \begin{subfigure}{0.19\textwidth}
        \centering
        \includegraphics[width=\linewidth]{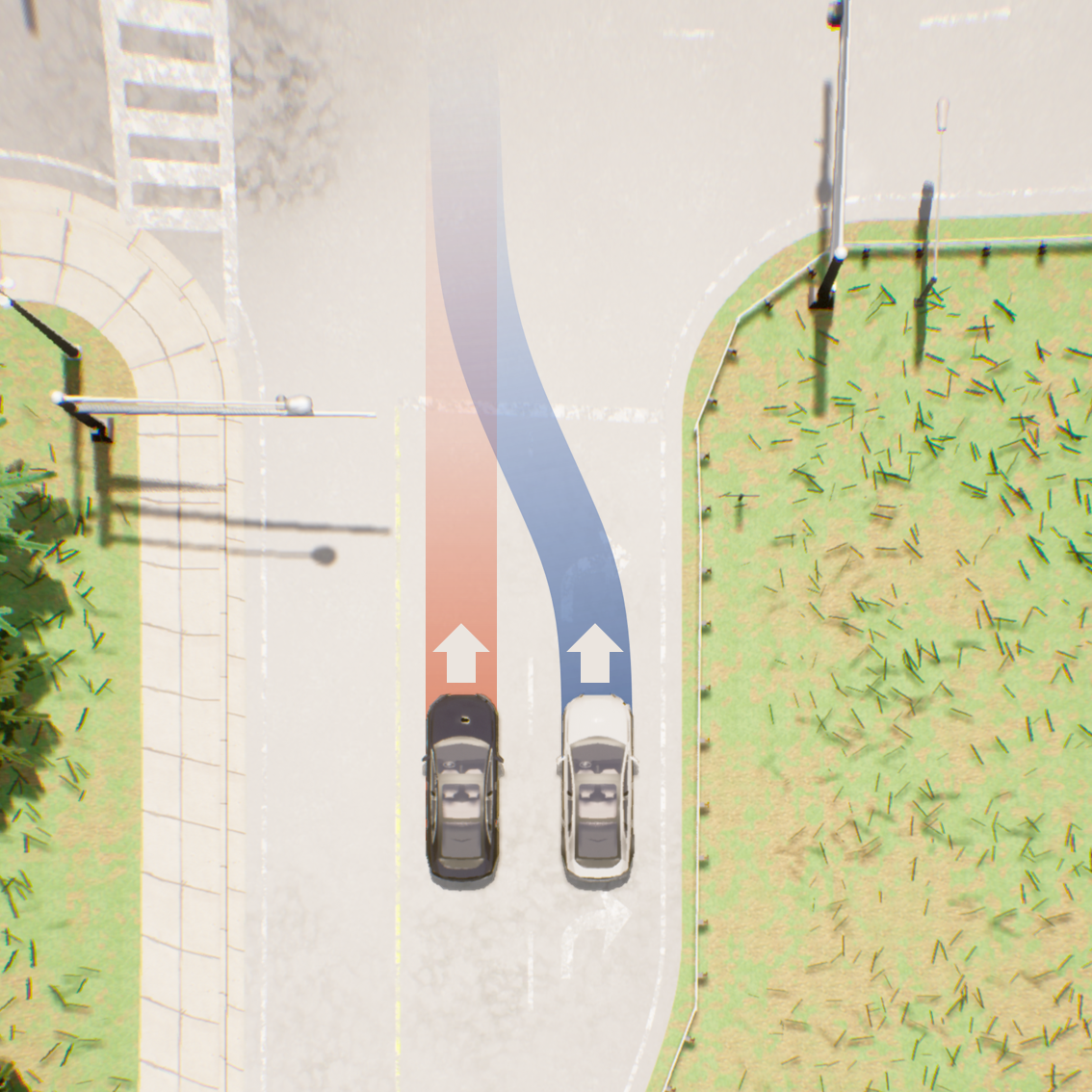}
        \caption{Merge:Neighbor Lane}
        \label{scen_6}
    \end{subfigure}  
    \begin{subfigure}{0.19\textwidth}
        \centering
        \includegraphics[width=\linewidth]{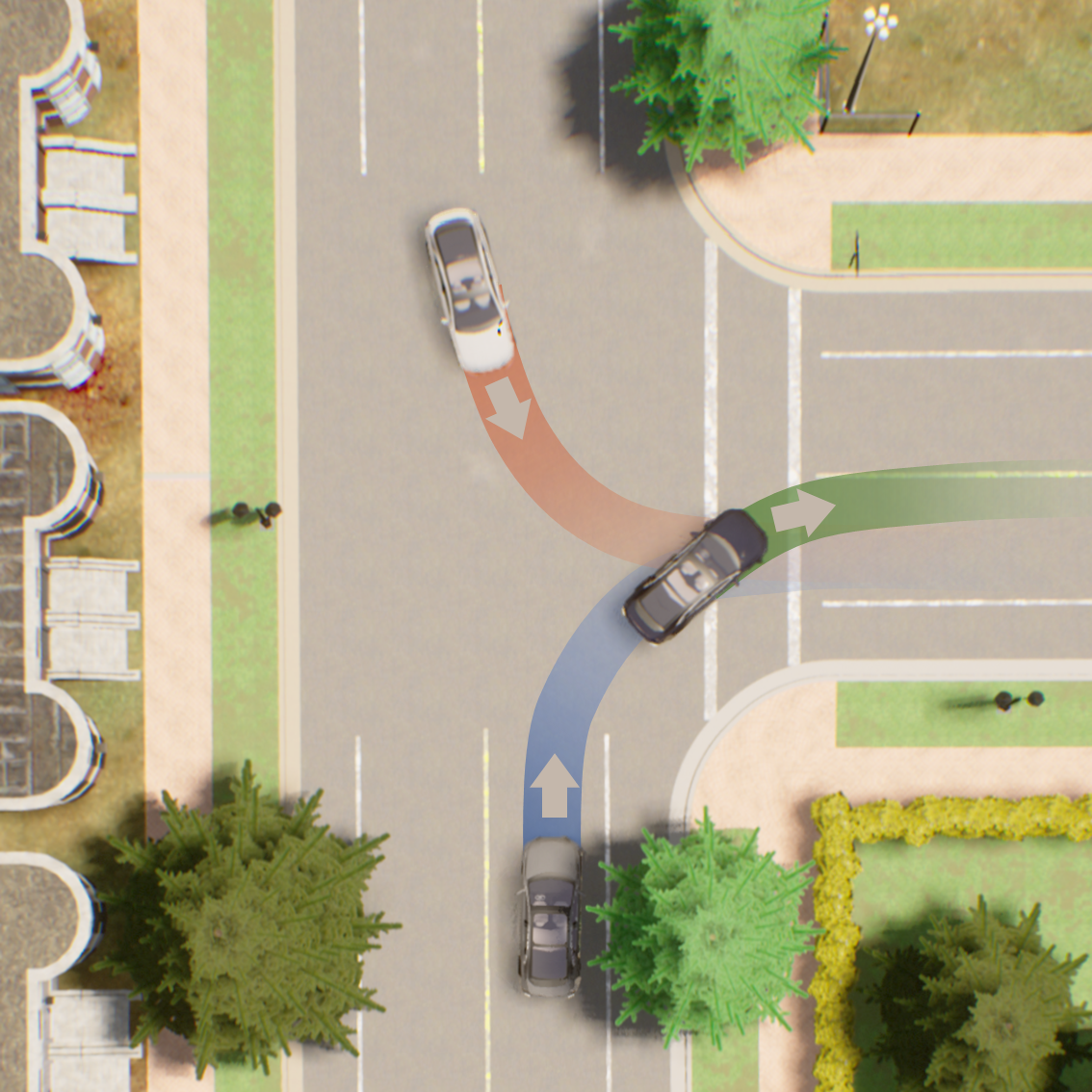}
        \caption{Merge:Left-Right}
        \label{scen_7}
    \end{subfigure}  
    \begin{subfigure}{0.19\textwidth}
        \centering
        \includegraphics[width=\linewidth]{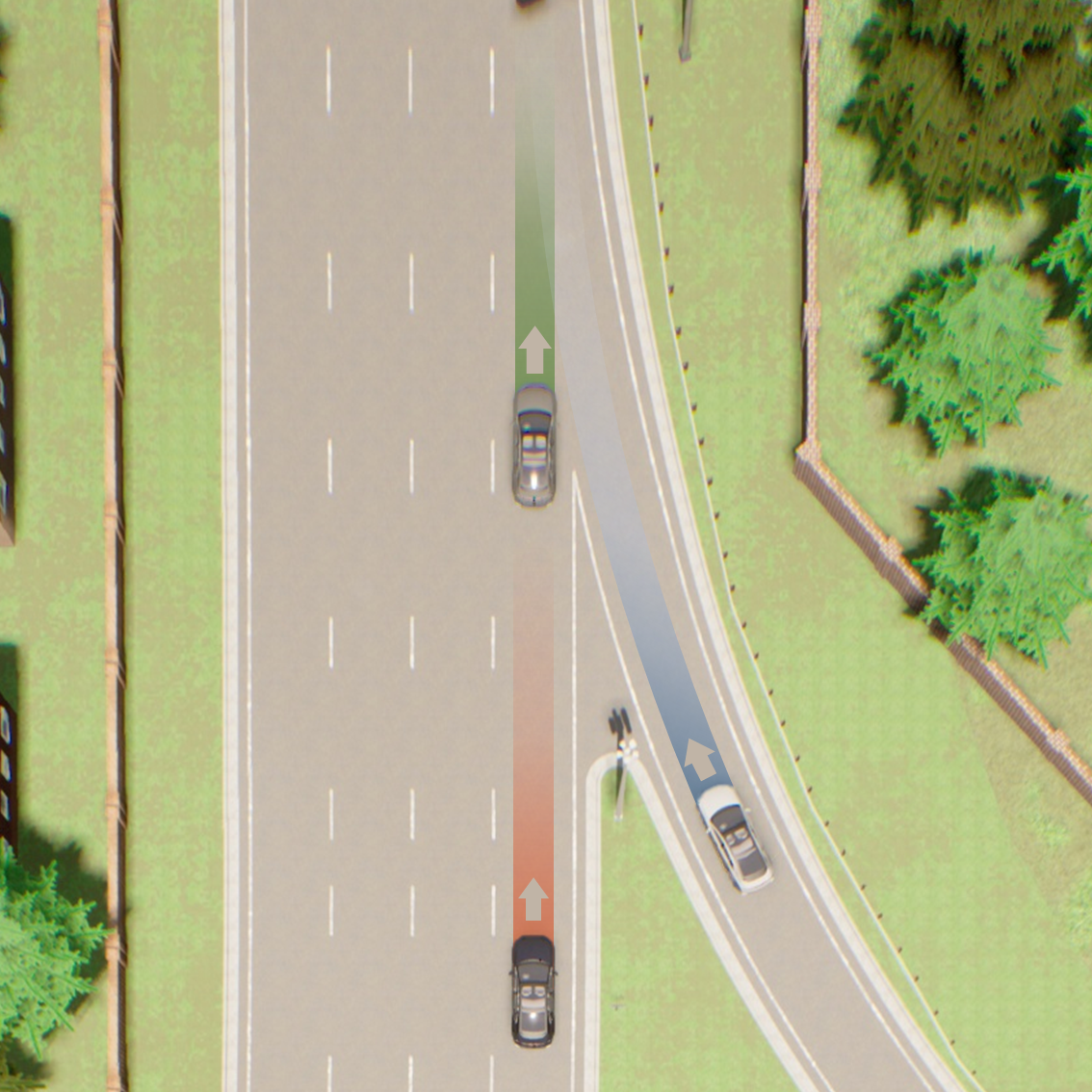}
        \caption{Merge:Highway}
        \label{scen_8}
    \end{subfigure}  
    \begin{subfigure}{0.19\textwidth}
        \centering
        \includegraphics[width=\linewidth]{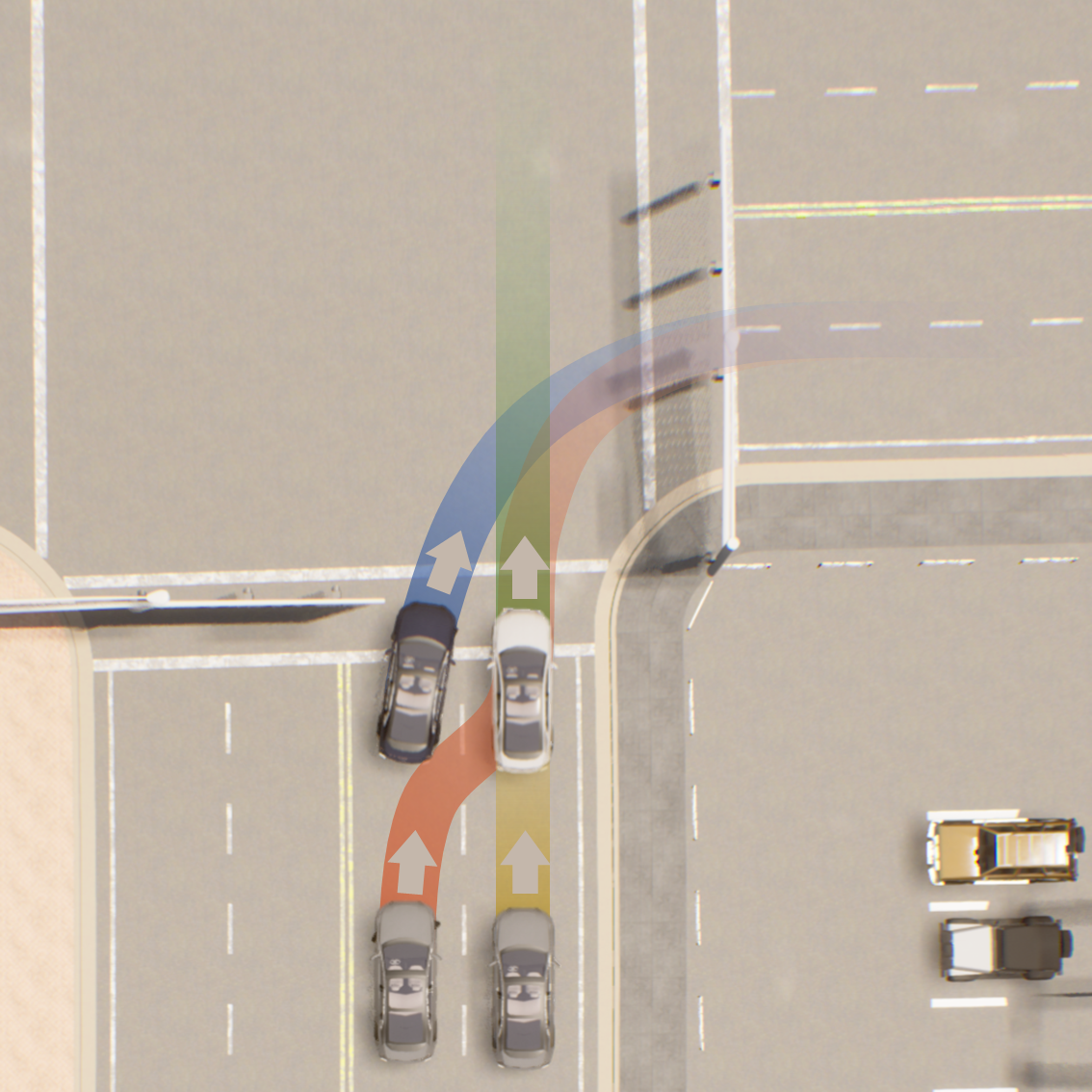}
        \caption{Change:Right-Straight}
        \label{scen_9}
    \end{subfigure}  
    \begin{subfigure}{0.19\textwidth}
        \centering
        \includegraphics[width=\linewidth]{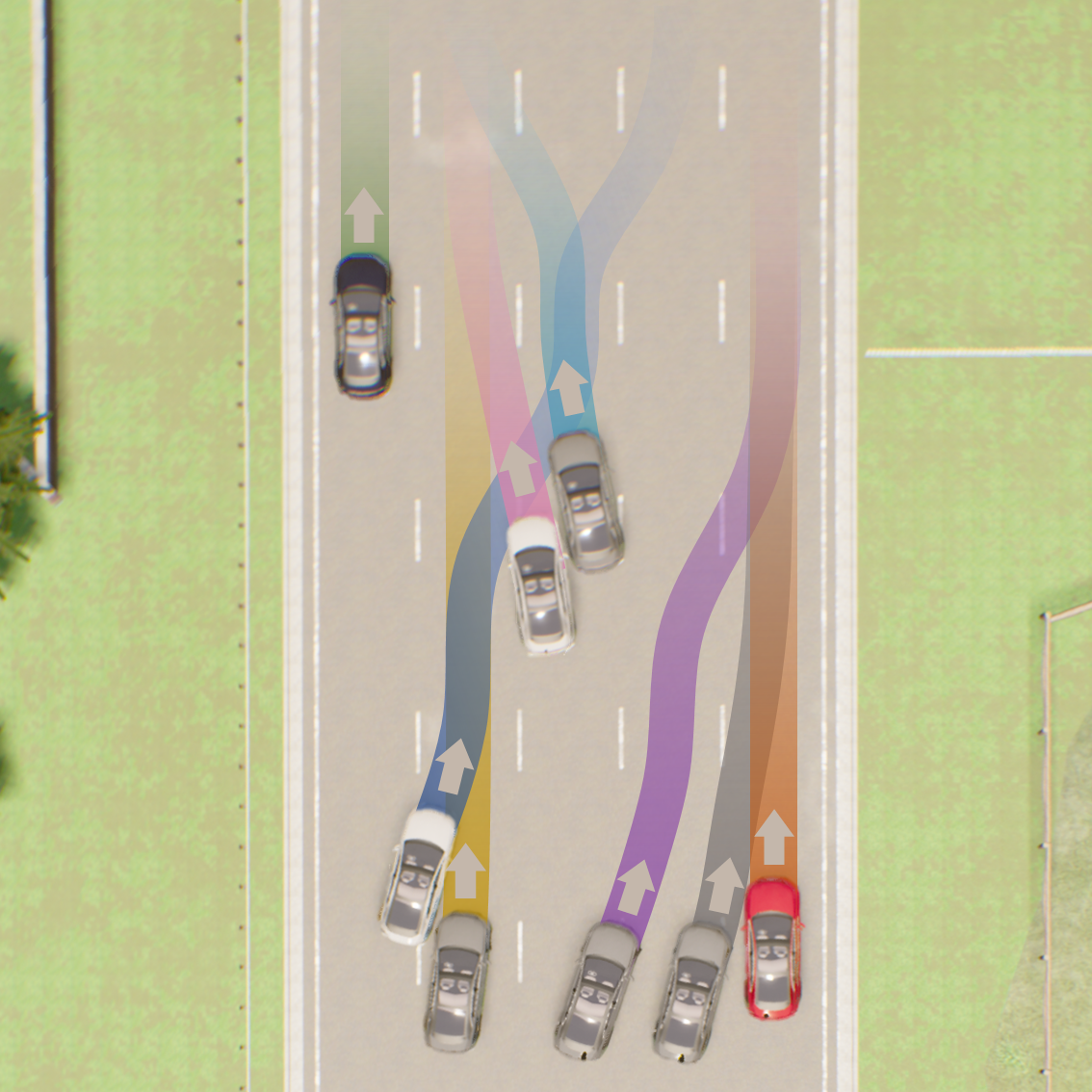}
        \caption{Change:Highway}
        \label{scen_10}
    \end{subfigure}  
    \caption{The 10 types of scenarios in the proposed InterDrive benchmark. These scenarios evaluate the three key skills in handling interaction among nearby vehicles, including going cross intersections (a-d), lane merging (e-h), and lane changing (i-j). }
    \vspace{-3mm}
    \label{fig:scenarios}
\end{figure*}

To evaluate the capabilities of autonomous driving systems in handling multi-vehicle interaction, we present the InterDrive benchmark on top of V2Xverse simulation platform. This benchmark encompasses 10 types of typical multi-vehicle scenarios, each involving multiple under-test vehicles. We assign these vehicles with largely overlapped target paths to encourage conflicts, and randomly deploy additional traffic participants (vehicles, pedestrians, cyclists) as obstacles. These scenarios are constructed to simulate realistic traffic scenarios where several on-road vehicles are autonomous-driven.



\vspace{-1mm}
\noindent
\subsection{Scenarios}
Fig~\ref{fig:scenarios} visualizes the 10 scenarios in InterDrive Benchmark, where we construct traffic scenarios with reference to the pre-crash typology of the US National Highway Traffic Safety Administration (NHTSA). Through these scenarios, we assess three typical scenarios in handling multi-vehicle interaction, including crossing intersections, lane merging, and lane changing.


\noindent
\textbf{*Intersection Crossing (IC).} Several vehicles enter, meet, and then exit an intersection from different directions. 
Four distinct types of scenarios are incorporated, with visual representations shown in Fig~\ref{fig:scenarios}(a)-(d). To ensure comprehensive evaluation diversity, we carefully design different combinations of entry and exit directions for the vehicles at the intersection.

\noindent
\textbf{*Lane Merging (LM).} Vehicles merge into the same lane from different directions, see visualizations in Fig~\ref{fig:scenarios}(e)-(h). We construct scenarios in different road topologies, including parallel straight-ahead lanes, T-junctions, and highway ramps.

\noindent
\textbf{*Lane Changing (LC).}
This study defined two distinct lane-changing scenarios. In these scenarios, multiple vehicles initially maintain parallel trajectories while traveling in the same direction. Subsequently, they are required to execute lane-changing maneuvers, intersecting the trajectories of adjacent vehicles to reach their respective destinations. See visualizations in Fig~\ref{fig:scenarios}(i),(j). 

\noindent
InterDrive benchmark extends each scenario through diverse configurations, varying in route waypoints, the number of vehicles under test, and additional obstacles, ultimately generating 92 distinct test tasks. The number of interactive test vehicles is configured to range from 2 to 8, simulating the typical number of vehicles with which a single vehicle may have direct conflicts simultaneously.


\subsection{Metrics}
InterDrive incorporates five metrics: \textit{Route Completion}, \textit{Infraction Score}, and \textit{Driving Score}, which are adopted from CARLA Leaderboard~\cite{carlaleaderboard}, along with a additional metrics: \textit{Success Rate} 

\noindent
\textit{Route Completion (RC)} is the percentage of the total planned route distance completed by the under-test vehicles.

\noindent
\textit{Infraction Score (IS)} starts at 1.0 for each task and reduced with collisions by a predefined discount factor, evaluating all test vehicles safety perfeomance.

\noindent
\textit{Driving Score (DS)} serves as the primary ranking metric, and is calculated as the product of Route Completion and Infraction Score, capturing both task progress and safety.

\noindent
\textit{Success Rate (SR)} is the percentage of tasks completed with a full-mark Driving Score, reflecting the consistency of the system to achieve reliable driving performance.


\noindent
These metrics collectively provide a comprehensive view of navigation performance in multi-vehicle driving scenarios.

\vspace{-2mm}

\begin{table*}[t!]
\centering
\caption{Driving performance in InterDrive Benchmark. CoLMDriver achieves the highest success rate in all scenarios.}
\label{tab:inter}
\resizebox{\textwidth}{!}{%
\begin{tabular}{c|cccc|cccc|cccc|cccc}
\toprule
\multirow{2}{*}{\textbf{Method}} & \multicolumn{4}{c|}{\textbf{InterDrive-total}} & \multicolumn{4}{c|}{\textbf{InterDrive-IC}} & \multicolumn{4}{c|}{\textbf{InterDrive-LM}} & \multicolumn{4}{c}{\textbf{InterDrive-LC}} \\
 & \textbf{DS↑} & \textbf{RC↑} & \textbf{IS↑} & \textbf{SR↑} & \textbf{DS↑} & \textbf{RC↑} & \textbf{IS↑} & \textbf{SR↑} & \textbf{DS↑} & \textbf{RC↑} & \textbf{IS↑} & \textbf{SR↑} & \textbf{DS↑} & \textbf{RC↑} & \textbf{IS↑} & \textbf{SR↑} \\ \midrule
VAD\cite{jiang2023vad} & 25.37 & 75.00 & 0.33 & 0.02 & 15.49 & 54.72 & 0.29 & 0.00 & 37.24 & 92.85 & 0.40 & 0.05 & 17.93 & 76.00 & 0.26 & 0.00 \\
UniAD\cite{UniAD} & 35.17 & 88.30 & 0.38 & 0.11 & 37.24 & 91.63 & 0.41 & 0.11 & 42.50 & 84.41 & 0.47 & 0.15 & 12.19 & 90.57 & 0.12 & 0.00 \\
TCP~\cite{TCP} & 73.68 & 90.54 & 0.82 & 0.50 & 77.64 & 82.83 & \textbf{0.94} & 0.50 & 82.18 & 95.18 & 0.86 & 0.70 & 43.52 & 96.30 & 0.45 & 0.00 \\
LMDrive~\cite{shao2024lmdrive} & 48.83 & 58.02 & 0.85 & 0.20 & 44.72 & 57.94 & 0.79 & 0.17 & 60.88 & 69.43 & 0.86 & 0.30 & 27.96 & 29.70 & \textbf{0.95} & 0.00 \\ \midrule
CoDriving~\cite{liu2024codriving} & 74.13 & \textbf{96.31} & 0.76 & 0.57 & 66.32 & 90.57 & 0.71 & 0.61 & 96.18 & \textbf{100.00} & 0.96 & 0.75 & 36.57 & \textbf{100.00} & 0.37 & 0.00 \\
Rule-based & 78.38 & 91.85 & 0.80 & 0.72 & 80.06 & \textbf{95.93} & 0.81 & \textbf{0.72} & 94.44 & \textbf{100.00} & 0.94 & 0.90 & 34.43 & 62.29 & 0.42 & 0.25 \\
CoLMDriver(Ours) & \textbf{88.53} & 94.05 & \textbf{0.90} & \textbf{0.80} & \textbf{82.07} & 88.78 & 0.86 & \textbf{0.72} & \textbf{98.27} & 99.93 & \textbf{0.98} & \textbf{0.92} & \textbf{59.21} & 82.50 & 0.597 & \textbf{0.50} \\ \bottomrule
\end{tabular}%
}
\end{table*}

\section{Experiments}

\begin{table}[]
\centering
\caption{Ablation study of system components. \textit{Nego} : negotiation, \textit{S/E} :safety/efficiency score, \textit{Cons}: consensus score.}
\label{tab:component}
\resizebox{\columnwidth}{!}{%
\begin{tabular}{l|cccc|cccc}
\toprule
\multirow{2}{*}{ID} & \multicolumn{1}{c|}{\multirow{2}{*}{\textbf{Nego}}} & \multicolumn{1}{c|}{\multirow{2}{*}{\textbf{Grouping}}} & \multicolumn{2}{c|}{\textbf{Critic}} & \multirow{2}{*}{\textbf{DS↑}} & \multirow{2}{*}{\textbf{RC↑}} & \multirow{2}{*}{\textbf{IS↑}} & \multirow{2}{*}{\textbf{SR↑}} \\
 & \multicolumn{1}{c|}{} & \multicolumn{1}{c|}{} & \textbf{S/E} & \textbf{Cons} &  &  &  &  \\ \midrule
1 &  &  &  &  & 47.64 & \textbf{96.43} & 0.485 & 0.130 \\
2 & \checkmark &  &  &  & 9.33 & 10.37 & 0.935 & 0.000 \\
3 & \checkmark & \checkmark &  &  & 77.29 & 95.22 & 0.784 & 0.652 \\
4 & \checkmark & \checkmark & \checkmark &  & 83.46 & 91.93 & 0.860 & 0.739 \\
5 & \checkmark & \checkmark & \checkmark & \checkmark & \textbf{88.53} & 94.05 & \textbf{0.903} & \textbf{0.804} \\ \bottomrule
\end{tabular}%
}
\end{table}

\subsection{Experimental Settings}
We implement and evaluate our method on the CARLA simulator of version 0.9.10.1~\cite{DosovitskiyCARLA:CoRL2017}. The simulation frequency is set to 5 Hz for all experiments except real-time ablation. For the low-level pipeline in the CoLMDriver framework, we deploy PointPillars~\cite{Lang2018PointPillarsFE} to encode point clouds. We use Lora finetuning~\cite{hu2021lora} for InternVL2-4B~\cite{chen2024internvl} as the VLM intention planner and Qwen2.5-3B~\cite{qwen2.5} as the LLM negotiator, for both accuracy and efficiency consideration. For the intention-waypoints translator, we use an embedding size of 256 and a medium feature size of 256 as well, with 20 output waypoints at 5 Hz.

\subsection{Quantitative Results of Closed-loop driving}

\noindent
\textbf{Performance in Highly Interaction Traffic Scenarios. }Tab.~\ref{tab:inter} presents CoLMDriver's driving performance in our proposed InterDrive Benchmark, compared to other advanced closed-loop driving baselines, including TCP~\cite{TCP}, VAD~\cite{jiang2023vad}, UniAD~\cite{UniAD}, CoDriving~\cite{liu2024codriving}, and another VLM-based method, LMDrive~\cite{shao2024lmdrive}. To prove the necessity of negotiation, we build the Rule-based method on traffic norms as comparison. Optimization-based cooperative planning methods are not compared due to being closed-source or on other platform. The table shows the overall score under InterDrive and separate performance for InterDrive-IC, InterDrive-LM and InterDrive-LC. CoLMDriver achieves SOTA performance on driving score(DS) across all interactive scenarios due to its language negotiation capability, outperforming other baselines by at least 10.15\% in DS. The three cooperative driving methods all outperform single-agent driving approaches, demonstrating the effectiveness of cooperation in conflict resolution. Other baselines face challenges such as target recognition issues, leading to lower route completion(RC), or collision incidents due to the lack of negotiation, resulting in low infraction scores(IS). TCP achieves a relatively high driving score but struggles with a low success rate (SR), indicating frequent collisions among scenes. LMDrive benefits from its multi-view, multimodal input and LLM capability, achieving a high infraction score, but encounters challenges in driving interruption where two cars come to a stop due to close proximity, each yielding to the other without progressing. Both intention-conflict collisions and dual-yielding issues can be resolved through language negotiation.

\begin{figure}[!t]
  \centering
  \includegraphics[width=0.45\textwidth]{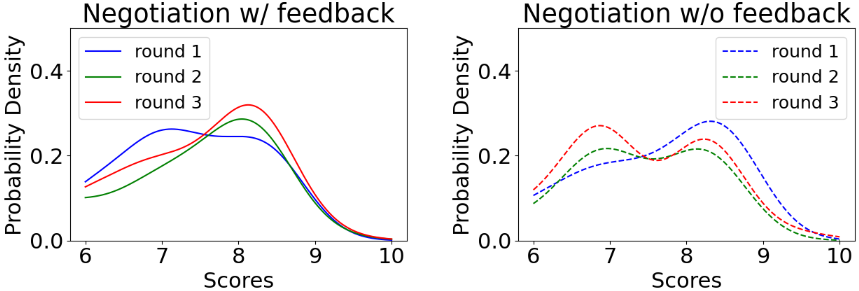}
  \caption{Experiment of negotiation convergence guided by Actor-Critic paradigm.}
  \vspace{-6mm}
  \label{fig:nego_conv}
\end{figure}
\noindent
\textbf{Consensus Convergence. }Fig.~\ref{fig:nego_conv} presents the negotiation quality score distribution of the evaluator for system with or without critic feedback. When the LLM updates its negotiation messages based on conversation alone, the negotiation quality score fluctuates randomly across rounds. However, when guided by evaluator feedback, the score exhibits a steady increase as negotiation iterates.

\noindent
\textbf{System Component Ablation. }Tab.~\ref{tab:component} evaluates the impact of different system components on performance. A system without negotiation (ID 1) performs closely to LMDrive on DS, demonstrating solid baseline performance. However, negotiation without the dynamic grouping mechanism leads to continuous stopping, resulting in lower route completion. Incorporating the Actor-Critic paradigm into the negotiation module further enhances the driving score.

\begin{figure}[!t]
  \centering
  \includegraphics[width=0.45\textwidth]{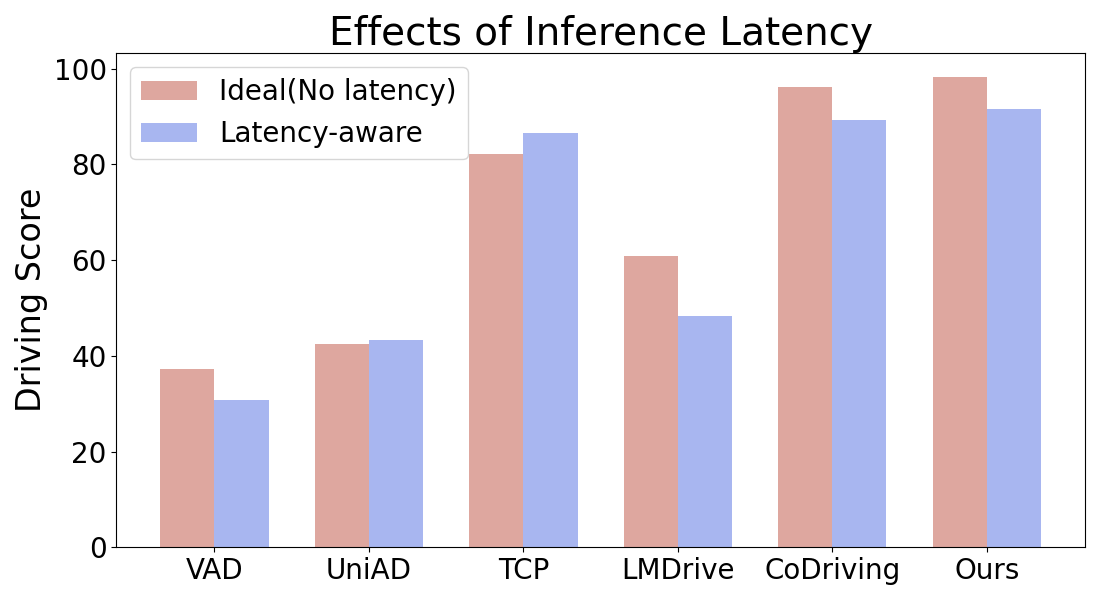}
  \caption{The driving performance with/without (Latency-aware/Ideal) accounting for inference latency.}
  \vspace{-6mm}
  \label{fig:realtime_ablation}
\end{figure}
\noindent
\textbf{Real-time Performance. }
We compare the performance in ideal computing situation(no inference latency) and situation with inference latency in Fig. \ref{fig:realtime_ablation} on the InterDrive-LM. Our CoLMDriver experiences only a 6.62\% drop in driving score and still keeps driving score over 90, demonstrating the inference efficiency of the proposed system. In our framework, the low-level planning pipeline can continuously generate precise execution based on intention guidance within varying environment. TCP, operating faster than our ideal simulation, slightly increase its performance.



\noindent
\textbf{Performance on public benchmark.} We further investigate the general navigation capability of CoLMDriver on the public Town05 benchmark~\cite{TransFuser}. To enable V2V communication in this single-vehicle driving benchmark, we enable the surrounding vehicles to transmit their driving intention to the ego vehicle but do not change their own behaviors. Tab.~\ref{tab:general} compares the driving CoLMDriver with two SOTA single-vehicle driving methods baseline methods. We can see that CoLMDriver achieves a superior Driving Score in both long and short routes, and surpasses ReasonNet by 11\% in Town05 Long. This is because CoLMDriver receives driving intention from neighbors, thereby reducing the uncertainty in planning.

\begin{table}[!h]
\vspace{-2mm}
\caption{Driving performance on Town05 benchmark~\cite{TransFuser}}
\centering
\label{tab:general}
\resizebox{0.4\textwidth}{!}{
\begin{tabular}{c|cc|cc}
\toprule
\multirow{2}{*}{Method} & \multicolumn{2}{c|}{Town05 Short} & \multicolumn{2}{c}{Town05 Long} \\
 & DS↑ & RC↑ & DS↑ & RC↑ \\ \midrule
InterFuser~\cite{InterFuser} & 94.95 & 95.19 & 68.31 & 94.97 \\
ReasonNet~\cite{shao2023reasonnet} & 95.71 & 96.23 & 73.22 & 95.88 \\
CoLMDriver(Ours) & \textbf{96.14} & \textbf{96.45} & \textbf{81.49} & \textbf{96.72} \\ \bottomrule
\end{tabular}}
\end{table}
\vspace{-3mm}

\section{Conclusion and limitation}
In this paper, we present CoLMDriver, an innovative autonomous driving system that leverages multimodal LLMs for effective language-based cooperative planning in end-to-end driving. CoLMDriver employs high-level driving intention to guide low-level waypoints generation, and utilizes multi-round negotiation to achieve consensus in highly interactive scenarios. Meanwhile, we construct the InterDrive Benchmark to evaluate autonomous driving systems in such interactive environments. Extensive closed-loop experiments demonstrate the effectiveness of CoLMDriver, highlighting the significant potential of language-based negotiation for advancing cooperative driving. 
One current limit is the diversity of language interaction demonstrations, which we aim to expand in future work by constructing more complex and interactive scenarios, further enhancing the system's capability and adaptability.

{
    \small
    \bibliographystyle{unsrt}
    \bibliography{main}
}


\clearpage
\setcounter{page}{1}
\maketitlesupplementary

\section{Model Details}

\subsection{VLM-based Intention Planner}
\noindent \textbf{Dataset and Training.} As described in Sec.~\ref{subsec:intention_planner}, we adopted a three-stage training approach for the VLM planner. In the first stage, Driving Knowledge Enhancement Training, we utilized a sampled DriveLM-CARLA dataset containing 64k image-QA pairs focused on driving-related knowledge for perception, prediction, and planning. This stage was completed in a single epoch. In the second stage, Intention Tuning, the VLM was fine-tuned on 50k samples of our collected intention dataset. For each frame, the input query was structured by incorporating the transformed GT perception information, GT navigation instructions and speed into the VLM prompt. The response combined navigation and speed intentions. Finally, in the third stage, Consensus Tuning, we enriched the second-stage dataset by adding negotiation information. The VLM was fine-tuned for five epochs in the second stage and one epoch in the third stage. Key training parameters included LoRA tuning, DeepSpeed ZeRO-3 optimization, a batch size of 1, and learning rates of \(1 \times 10^{-4}\) for stages one and two, and \(1 \times 10^{-5}\) for stage three. For reference, we trained the InternVL2-4B model on 8 NVIDIA 3090 GPUs, with each epoch taking approximately 5 hours.

\subsection{Intention-guided Waypoint Planner}
\noindent \textbf{Model Structure.} The Occupancy Encoder and the Feature Encoder each comprise two convolutional layers with 32 output channels. The MotionNet Encoder includes two Spatial-Temporal Convolution blocks followed by a standard convolutional block. Each Spatial-Temporal block consists of two 2D convolutional layers and one 3D convolutional layer. The MotionNet Encoder generates an output with 256 channels. Both the speed intention and direction intention are embedded into 256-dimensional vectors, matching the output channels of the MotionNet Encoder. Similarly, the target point is transformed into a 256-dimensional vector using a three-layer MLP. Then the MLP Fuser combines the concatenated vector into 256 dimensions. The Transformer Decoder, which includes three layers, applies cross-attention and self-attention mechanisms to BEV Tokens and Command Tokens. Finally, a two-layer MLP decoder predicts 10 waypoints, which are used as control signals.

\noindent \textbf{Dataset.} The dataset used for training and testing the generator is derived from CARLA. The training set is constructed from data in CARLA towns 1, 2, 3, 4, and 6, while data from towns 7, 8, and 10 are used for validation, and town 5 is reserved for testing. The original training dataset consists of approximately 25k records. We extend the dataset into four categories—STOP, SLOWER, KEEP, and FASTER. This is done by first polynomial fitting the original trajectory and then sample waypoints according to the intention and environmental information. Then the actual training dataset grows to approximately 93k records. This number is slightly less than four times the original dataset size, as in certain cases, the original trajectory is too short for polynomial fitting. The perception module processes the last five frames of data, and outputs BEV features and BEV occupancy. The BEV occupancy contains six channels with a resolution of  $192\times 96$, while the BEV feature comprises 128 channels at the same resolution. Intentions are represented as indexing tensors corresponding to the category of the given extended record. Training the generator on 8 NVIDIA 3090 GPUs takes approximately 9 hours per epoch, with convergence typically achieved after 10 epochs.

\section{InterDrive Benchmark Details}
In the current iteration of the InterDrive Benchmark, we have meticulously selected 46 routes from the Town05, Town06, and Town07 scenarios within the CARLA simulator. 

\noindent
\textbf{Route distribution.}
In the InterDrive Benchmark, which consists of 46 routes, 10 scenario types and 3 categories. We integrate the characteristics of the scenarios with the specific conditions of each Carla Town to ensure that each route is both challenging to complete and practically valuable. Town05, characterized by an urban environment, is the most representative of the model's target application environment, hence its higher number of routes. Town06 is distinguished by its multi-lane highways, whereas Town07 primarily features rural scenarios with narrow roads. We have designed a variety of scenarios by varying the number of vehicles and the surrounding environments, which include different towns and diverse intersections, as shown in Tab. \ref{tab:route_distribution}. Their inclusion in the benchmark is crucial for enhancing its diversity and significantly raises the complexity of driving tasks, particularly in terms of vehicle-vehicle interactions.

\begin{table}[!t]
\centering
\footnotesize
\caption{Detailed information of the 10 scenario types in InterDrive Benchmark.}
\label{tab:route_distribution}
\begin{tabular}{ccccc}
\toprule
\makecell{\textbf{Scenario}\\ \textbf{Type}} & \makecell{\textbf{Scenario}\\ \textbf{Category}} & \makecell{\textbf{Vehicle}\\ \textbf{Count}} & \makecell{\textbf{Carla}\\ \textbf{Town No.}} & \makecell{\textbf{Route}\\ \textbf{Count}}\\ \midrule
Straight-Straight & IC & 2 & 05, 06, 07 &4\\ 
Straight-Left & IC & 2 & 05, 06, 07 &6\\ 
Opposite Lane & IC & 3, 4 & 05 &4\\ 
Chaos & IC & 6, 8 & 05 &4\\ 
Straight-Right & LM & 2 & 05, 06, 07 &6\\ 
Neighbor Lane & LM & 2 & 05, 06, 07 &6\\ 
Left-Right & LM & 3, 4 & 05 &4\\ 
Highway-Merge & LM & 3, 4 & 06 &4\\ 
Right-Straight & LC & 3, 4 & 05 &4\\ 
Highway-Change & LC & 6, 7, 8 & 06 &4\\ 
\bottomrule
\end{tabular}
\end{table}

\noindent
\textbf{Scenario Settings.}
In order to enhance the fidelity of the simulation environment to real-world scenarios, we have introduced a certain number of additional traffic participants. Specifically, we set the number of vehicles, pedestrians, and cyclists in the environment to 50 each. This allows them to create a certain level of disturbance without completely blocking the routes and interfering with the predefined vehicle interactions. Moreover, this numerical value is also objectively close to the actual traffic conditions in real-world scenarios.

The result of the simulation with these participants are shown in Tab.~\ref{tab:npc}. By comparing with Tab.~\ref{tab:inter}, it can be observed that the inclusion of traffic participants has a certain impact on the methods primarily based on cooperation. In contrast, the scores of non-cooperative methods remain essentially unchanged or even slightly improve. This is because these participants still prevent the originally designed vehicle conflicts in specific scenarios.

\begin{table*}[!t]
\centering
\caption{Driving performance in InterDrive Benchmark with traffic participants.}
\label{tab:npc}
\resizebox{\textwidth}{!}{%
\begin{tabular}{c|cccc|cccc|cccc|cccc}
\hline
\multirow{2}{*}{\textbf{Method}} & \multicolumn{4}{c|}{\textbf{InterDrive-total}} & \multicolumn{4}{c|}{\textbf{InterDrive-IC}} & \multicolumn{4}{c|}{\textbf{InterDrive-LM}} & \multicolumn{4}{c}{\textbf{InterDrive-LC}} \\
 & \textbf{DS↑} & \textbf{RC↑} & \textbf{IS↑} & \textbf{SR↑} & \textbf{DS↑} & \textbf{RC↑} & \textbf{IS↑} & \textbf{SR↑} & \textbf{DS↑} & \textbf{RC↑} & \textbf{IS↑} & \textbf{SR↑} & \textbf{DS↑} & \textbf{RC↑} & \textbf{IS↑} & \textbf{SR↑} \\ \hline
VAD & 25.18 & 75.66 & 0.31 & 0.02 & 22.13 & 61.31 & 0.32 & 0.00 & 35.10 & 88.23 & 0.39 & 0.05 & 7.24 & 76.56 & 0.09 & 0.00 \\
UniAD & 37.13 & 88.71 & 0.41 & 0.11 & 37.45 & 83.52 & 0.44 & 0.06 & 48.29 & 91.33 & 0.52 & 0.20 & 8.48 & 93.82 & 0.09 & 0.00 \\
TCP & 74.18 & 91.21 & 0.82 & 0.48 &  \textbf{76.26} & 84.62 & \textbf{0.91} & 0.44 &  86.59 & 95.00 & 0.91 & 0.65 &  38.50 & 96.56 & 0.40 & 0.13  \\
LMDrive & 49.95 & 61.61 & \textbf{0.84} & 0.13 &  47.65 & 59.34 & 0.81 & 0.00 &  54.12 & 67.10 & 0.85 & 0.20 &  44.69 & 53.02 & \textbf{0.87} & 0.25  \\ \hline
CoDriving & 64.50 & \textbf{93.58} & 0.67 & 0.47 & 54.64 & 88.08 & 0.59 & 0.33  & 88.07 & 96.55 & 0.91 & 0.73 &  27.78 & \textbf{98.51} & 0.28 & 0.13  \\
Rule-based & 69.71 & 87.35 & 0.75 & 0.57 & 66.05 & \textbf{88.48} & 0.72 & \textbf{0.50} & 90.72 & 97.92 & 0.93 & 0.80 & 25.44 & 58.38 & 0.38 & 0.13 \\
CoLMDriver & \textbf{77.09} & 92.02 & 0.80 & \textbf{0.63}  & 63.06 & 82.55 & 0.70 & 0.44  &  \textbf{94.00} & \textbf{100.00} & \textbf{0.94} & \textbf{0.85} &  \textbf{66.38} & 93.41 & 0.68 & \textbf{0.50 } \\ \hline
\end{tabular}%
}
\end{table*}

\section{Prompt Details and Example}

\lstset{
    basicstyle=\ttfamily\footnotesize, 
    backgroundcolor=\color{gray!10}, 
    frame=single, 
    breaklines=true, 
    numbers=none, 
    captionpos=b, 
    breakindent=0pt,                   
}

\subsection{Prompt for VLM}
To better harness the knowledge and reasoning capabilities of the VLM, as well as to standardize its output format, we designed the VLM prompt based on the following structure: role definition, task description, logical guidance, additional rules, real-time input, and output format. The specific prompt design is detailed in Lst. \ref{lst: vlm_prompt}. The content in '\{\}' will be replaced by real-time input.

\begin{figure*}[ht]
\centering
\begin{lstlisting}[caption={VLM intention generation prompt}, label={lst: vlm_prompt}]
Suppose you are an autopilot assistant driving a car on the road. You will receive images from the car's front camera and are expected to provide driving intentions. There are other traffic participants in the scenario, and you may have communication with them. Your analysis logic chain should be as follows:
1. Understand the direction of the road and your own position.
2. Perceive surrounding objects.
3. Pay attention to key objects and dangerous situations.
4. Follow the rules listed below.
5. Check communication decision.
6. Finally, conclude the situation and provide driving intentions.

### Rules
1. If the environment is safe and clear, drive fast
2. Maintain a safe distance from the car in front.
3. Stop to avoid pedestrians preparing to cross the road.
4. Slow down or stop when other vehicles change lanes, merge or turn.
5. Slow down or stop when there is obstacle on the road ahead.
6. When establishing communication with other vehicles, take the communication decision as important reference.

### Perception Results
{perception}
### Real-time Inputs
Negotiation suggestion: {negotiation message}
Target direction: {navigation instruction}
Current Speed: {speed} m/s

### Output Requirements
Provide the navigation and speed intention. Navigation intention include 'turn left at intersection', 'turn right at intersection', 'go straight at intersection', 'follow the lane', 'left lane change', 'right lane change'. Speed intention include STOP, FASTER, SLOWER, KEEP.
\end{lstlisting}
\end{figure*}

\subsection{Prompt for LLM}
According to the design of our negotiation module, the prompts designed for the LLM consist of three types: vehicle-to-vehicle communication, sum actions, and consensus scoring.

In each round of negotiation, each vehicle broadcasts messages based on the prompts required for communication, and subsequently, one vehicle acts as a critic to sum actions and score. The prompts required for vehicle-to-vehicle communication are shown at Lst.~\ref{lst: llm_prompt_0}, where the environmental information and message records are denoted by '\{\}' and will change in real-time based on the scenario. The prompts for action-summing are presented in Lst.~\ref{lst: llm_prompt_1}, with the output being a JSON-formatted behavior request. The consensus scoring is then conducted using the prompts designed for evaluation, as shown in Lst.~\ref{lst: llm_prompt_2}, to complete one round of negotiation. Herein, the placeholder '-conv-' will be dynamically replaced with the current message record.

\begin{figure*}[ht]
\centering
\begin{lstlisting}[caption={LLM negotiation prompt - ego vehicle communication}, label={lst: llm_prompt_0}]
## Role
You are a driving assistant of a car (Vehicle ID: {i}). Given a scenario where multiple vehicles are in conflict, you need to negotiate with other vehicles to reach a consensus and ensure the safety and efficiency of all vehicles involved.

## Scenario
- Ego Vehicle (ID: {info['ego_id']}): Intention = {info['ego_intention']}, Speed = {round(info['ego_speed'], 1)}m/s
- Surrounding Vehicles:
{veh_string}

## Traffic Rules
0. In emergency situations, allow vehicles with special circumstances to pass through first.
1. Merging cars slow down to yield to straight car.
2. Left-turn cars slow down to yield to straight/right-turn car.
3. The car being yielded to go faster.
4. Cars behind decrease speed during emergency braking.
5. Following cars maintain a safe distance.

## Task
Based on the scenario info and conversation history, analyze the situation considering the **speed, direction, distance and intention of each vehicle**. Make sure you understand the situation before making any decisions. Pay attention to the traffic rules and critic suggestion. Identify any potential conflicts and propose actions that ensure the safety and efficiency of all vehicles involved. Remember to consider others' actions and requests from previous conversations. When conflicts occur, either request others to yield or yield to others. 
Your message may contain the action you will take and requests for other vehicles. **The actions and requests are speed intentions**

## Negotiation Tips
- Your actions should be logically consistent with your requests. No need for both sides to yield.
- Clearly specify which vehicle is responsible for each request or action.
- Focus your message on speed rather than navigation.

## Conversation History
{previous_conv}{sug_str}

## Output
You are vehicle {info['ego_id']}, you need to send a message to other cars. Please output the message only, within 18 words. Please do not provide specific speed values; instead, describe the trend of speed changes.
Sample output: I will [speed intention]; [requested speed intention].
\end{lstlisting}
\end{figure*}

\begin{figure*}[ht]
\centering
\begin{lstlisting}[caption={LLM negotiation prompt - sum actions}, label={lst: llm_prompt_1}]
## Task
Given a conversation of multiple cars negotiating to reach consensus, classify each vehicle's speed change into [STOP, SLOWER, KEEP, FASTER] and output the result as a string in format: {'id': car_id, 'speed': category}.

## Classification rules
- STOP: Come to a complete stop.
- SLOWER: Decrease speed.
- KEEP: Maintain current speed.
- FASTER: Increase speed.

## Additional rules:
- If a car request others to yield, it should go faster
- If a car yield to others, it should stop

## Input conversation:
{conv}
Your task is to analyze the given conversations for each vehicle and output the classification as a string in the specified format. DO NOT output other content other than the required actions. Ensure the output matches the required structure exactly.

## Output example:
{"0": {"speed": "STOP"}, "1":{"speed": "SLOWER"}, "2":{"speed": "SLOWER"}...}
\end{lstlisting}
\end{figure*}

\begin{figure*}[ht]
\centering
\begin{lstlisting}[caption={LLM negotiation prompt - consensus score evaluation}, label={lst: llm_prompt_2}]
Task Description:
Please analyze the following conversation and determine whether the characters have reached a consensus in the given scenario. Your response should include two parts: the first part is a brief explanation of whether a consensus was reached; the second part is a score indicating the degree of consensus, ranging from 0 to 100, where 0 means no consensus at all, and 100 means complete consensus.

Scoring Criteria:
0-20: There are significant disagreements with almost no common ground.
21-40: While there are some disagreements, there are one or two points where both parties can accept each other's views.
41-60: There is a moderate level of compromise and understanding on most discussed topics, but important disagreements remain unresolved.
61-80: Consensus has been reached on most issues, with only minor differences of opinion on a few details.
81-100: Almost all issues have been agreed upon by all parties, with only negligible objections remaining.

Scenario: On the road, multiple cars may have driving conflicts now. They negotiate with each other to avoid conflict.
Conversation:
{conv}

Your output format:
Short analysis: very short sentence to sum the consensus situation of the conversation.
Consensus score: int
\end{lstlisting}
\end{figure*}

\section{Autonomous Vehicle Details}
The autonomous vehicle in CoLMDriver processes sensor data and produces control signals as its final output. This section offers a detailed introduction to the sensor setup and controller configuration.

\noindent
\textbf{Sensor configurations.}
In CoLMDriver, we use the front-facing image with a resolution of 3000$\times$1500 and a horizontal field of view (FoV) of $100^\circ$ as an input for the VLM-based intention planner. For 3D detection, we rely on point clouds generated by a 64-channel LiDAR mounted at a height of 1.9 meters, with an upper FoV of $10^\circ$ and a lower FoV of $-30^\circ$.

\noindent
\textbf{Controller configurations.}
The controller generates executable driving actions, including steering, throttle, and braking, based on the predicted waypoints. To achieve this, we employ two PID controllers: a lateral controller and a longitudinal controller, which produce the corresponding control signals.
The lateral signal (turn signal) is calculated using the angle between the last two predicted waypoints, while the longitudinal signal (speed signal) is calculated using the average displacement in the predicted waypoints. Subsequently, we use the PID controller to generate a relatively smooth output. Mathematically, let $E \in \mathbb{R}^{N}$ be the historical signal with a time length of $N$, each PID controller takes the current signal $x$ as input and outputs 
$
x' = K_P * x + K_I * \mathrm{MEAN} (E) + K_D * (E[-1] - E[-2])
$
, where [$K_P$, $K_I$, $K_D$, N] forms a set of hyper-parameters for a PID controller. Specifically, the lateral controller is configured with [1, 0.2, 0.1, 5], while the longitudinal controller uses [5, 1, 0.1, 20].

\end{document}